\documentclass[10pt,journal,nodraftcls,compsoc]{IEEEtran}
\ifCLASSOPTIONcompsoc
  \usepackage[nocompress]{cite}
  \usepackage{times}
\usepackage{epsfig}
\usepackage{graphicx}
\usepackage{amsmath}
\usepackage{amssymb}
\usepackage{booktabs, multicol, multirow}
\usepackage{array}
\usepackage{xcolor}
\usepackage{hyperref}
\usepackage{subfigure}
\usepackage{float}
\usepackage{epstopdf}
\usepackage{mathtools}
\usepackage{setspace}
\usepackage{balance}
\usepackage{soul}

\usepackage[export]{adjustbox}

\DeclarePairedDelimiter\floor{\lfloor}{\rfloor}
\else
  \usepackage{cite}
\fi
\ifCLASSINFOpdf
\else
\fi
\newcommand{\etal}{\textit{et al}.}

\begin{document}
%
\title{Error-Corrected Margin-Based Deep Cross-Modal Hashing for Facial Image Retrieval}
%
%
%
%

\author{Fariborz Taherkhani *\thanks{* Authors Contributed Equally},~\IEEEmembership{Student Member,~IEEE,}
        Veeru Talreja * ,~\IEEEmembership{Student Member,~IEEE,}\\ Matthew C. 
        Valenti,~\IEEEmembership{Fellow,~IEEE,}
        and~ Nasser M. Nasrabadi,~\IEEEmembership{Fellow,~IEEE}}
\markboth{IEEE TRANSACTIONS ON BIOMETRICS, BEHAVIOR, AND IDENTITY SCIENCE}%
{Shell \MakeLowercase{\textit{et al.}}: Bare Demo of IEEEtran.cls for IEEE Journals}
%



\IEEEtitleabstractindextext{%
\begin{abstract}
Cross-modal hashing facilitates mapping of heterogeneous multimedia data into a common Hamming space, which can be utilized for fast and flexible  retrieval across different modalities. In this paper, we propose a novel cross-modal hashing architecture-deep neural decoder cross-modal hashing (DNDCMH), which uses a binary vector specifying the presence of certain facial attributes as an input query to retrieve relevant face images from a database. The DNDCMH network consists of two separate components: an attribute-based deep cross-modal hashing (ADCMH) module, which uses a margin (m)-based loss function to efficiently learn compact binary codes to preserve similarity between modalities in the Hamming space, and a neural error correcting decoder (NECD), which is an error correcting decoder implemented with a neural network. The goal of  NECD network  in DNDCMH  is to error correct the hash codes generated by ADCMH to improve the retrieval efficiency. The NECD network is trained such that it has an error correcting capability greater than or equal to the margin (m) of the margin-based loss function. This results in NECD can correct the corrupted hash codes generated by ADCMH up to the Hamming distance of m.  We have evaluated and compared DNDCMH with state-of-the-art cross-modal hashing methods on standard datasets  to  demonstrate the superiority of our method.

\end{abstract}

}

\maketitle

\IEEEdisplaynontitleabstractindextext

%
\IEEEpeerreviewmaketitle

\IEEEraisesectionheading{\section{Introduction}\label{sec:introduction}}

\IEEEPARstart{D}{ue} to the rapid development of the Internet and increasing usage of social media over the last decade, there has been a tremendous volume of multimedia data, which is generated from different heterogeneous sources and includes modalities like images, videos, and text. The approximate nearest neighbor (ANN) search has attracted significant attention from machine learning and computer vision  communities as it guarantees retrieval quality and computing efficiency for content based image retrieval (CBIR) in large-scale multimedia datasets. As a fast and advantageous solution, hashing has been employed in ANN search for CBIR due to its fast query speed and low storage cost \cite{angular_gong_2012,iterative_gong_2013, erin2015deep, lu2017deep,8691805,8931635,jiang2019discrete,8478207,8846593,8543225}. The goal of hashing is to map  high-dimensional visual data  to compact binary codes in Hamming space, where the semantic similarity in the original space is approximately preserved in Hamming space. The key principle in hashing functions is to maintain the semantic similarity by mapping images of similar content to similar binary codes. 

\begin{figure}[t] 
\centering
\includegraphics[width=8.8cm]{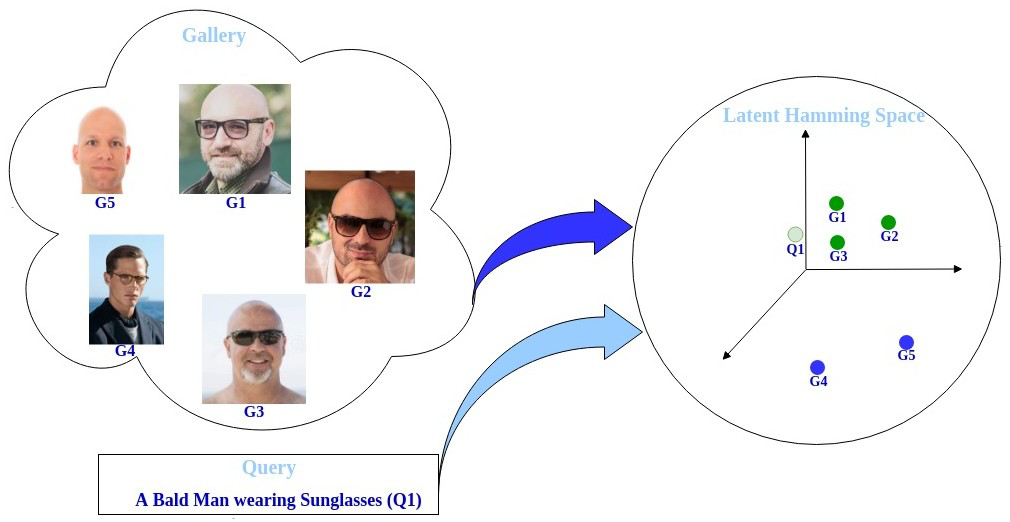}
\caption{Cross modal hashing for  facial image retrieval: a  bald  man  wearing Sunglass. }\label{fig:cmh}
\vspace{-0.20cm}
\end{figure}

Additionally, corresponding data samples from heterogeneous modalities may establish semantic correlations, which leads to cross-modal hashing (CMH). CMH returns relevant information of one modality in response to a query of another modality (e.g., retrieval of texts/images by using a query image/text), where similar hash codes in a shared latent Hamming space are generated for each individual modality. In this paper, we utilize the cross-modal hashing framework for a facial retrieval biometrics application in which  the images are retrieved  based solely on semantic attributes. For example, a user can give a query such as ``A bald man wearing sunglasses" to retrieve relevant face images from a large-scale gallery. The idea of cross-modal hashing for image retrieval applications is shown in Fig. \ref{fig:cmh}. We can note that the gallery's relevant points G1, G2 and G3 are closer to the query Q1 in the latent Hamming space than the points G4 and G5.

There has been a surge in the development of CMH techniques used for ANN search for retrieval on multi-modal datasets. However, capturing the semantic correlation between the heterogeneous data from divergent modalities \cite{Lin_SePH_2015}, and bridging the semantic gap between low-level features and high-level semantics for an effective CMH is a challenge. 
Deep learning techniques for CMH (or ``deep cross-modal hashing (DCMH)") \cite{Cao_THN_2017,jiang2017deep,Yang_2017_Pairwise,duan2018graphbit,8827941,8423193,8970562,8954946}  integrate feature learning and hash coding into an end-to-end trainable framework. DCMH frameworks minimize the quantization error of hashing from continuous representation to binary codes and provide the benefit of jointly learning the semantic similarity preserving features. 

The main goal in DCMH is to learn a set of hash codes such that the content similarities between different modalities is preserved in Hamming space. As such, a likelihood function \cite{jiang2017deep,chen2018dual,liong2017cross} or margin-based loss function such as the triplet loss function \cite{deng2018triplet,cao2016deep} needs to be incorporated into the DCMH framework to improve retrieval performance. In triplet-based DCMH \cite{deng2018triplet}, the  inter-modal triplet embedding loss encourages the heterogeneous correlation across
different modalities, and the intra-modal triplet loss encodes the discriminative power of the hash codes. Moreover, a regularization loss is used to apply \emph{adjacency consistency} to ensure that the hash codes can keep the original similarities in Hamming space. However,  in margin-based loss functions, some of the instances of different modalities of the same subject may not be close enough in Hamming space to guarantee all the correct retrievals. Therefore, it is important to bring the different modalities of the same subject closer to each other in Hamming space to improve the retrieval efficiency.

In this work, we observe  that in addition to the regular DCMH techniques \cite{jiang2017deep, cao2017collective, wang2016comprehensive}, which exploit entropy maximization and quantization losses in the objective function of the DCMH, an error-correcting code (ECC) decoder can be used as an additional component to compensate for the heterogeneity gap and reduce the Hamming distance between the different modalities of the same subject in order to improve the cross-modal retrieval efficiency. We presume that the hash code generated by DCMH is a binary vector that is within a certain distance from a codeword of an ECC. When the hash code generated by DCMH is passed through an ECC decoder, the closest codeword to this hash code is found, which can be used as a final hash code for the retrieval process. In this process, the attribute hash code and image hash code of the same subject are forced to map to the same codeword, thereby reducing the distance of the corresponding hash codes. This brings more relevant facial images from the gallery closer to the attribute query, which leads to an improved retrieval performance.

Recent work has shown that the same kinds of neural network architectures used for classification can also be used to decode ECC codes \cite{2016_nachmani_NND,nachmani2017rnn,Lugosch_2017_NeuralOM}. Motivated by this, we have used a neural error-correction decoder (NECD) \cite{2016_nachmani_NND} as an ECC decoder to improve the cross-modal retrieval efficiency. The NECD is a non-fully connected neural network architecture based on the belief propagation (BP) algorithm, which is a notable decoding technique applied for error-correcting codes. We have equipped  our ``attribute-based deep cross-modal hashing (ADCMH)" with the NECD to formulate our novel  deep neural decoder cross-modal hashing (DNDCMH) framework for cross-modal retrieval (face image retrieval based on semantic attributes), which, as we will demonstrate, performs better than other state-of-the-art deep cross-modal hashing methods for facial image retrieval.

Specifically, the DNDCMH  contains a custom-designed ADCMH network integrated with the NECD. The goal of ADCMH network is to learn pairwise optimized intermediate hash codes for both modalities, while the goal of NECD is to refine the intermediate hash codes generated by ADCMH  to improve the cross-modal retrieval efficiency.  The entire DNDCMH network is trained end-to-end  by implementing an alternative minimization algorithm in two stages. Stage 1 is split into parts 1(a) and 1(b). In Stage 1(a), ADCMH is trained by using a novel cross-modal loss function that uses a margin-based distance logistic loss (DLL). Stage 1(a) of the algorithm generates intermediate cross-modal hash codes. In stage 1(b), a NECD network is trained by relating the error correcting capability $e$ of the ECC used to create the NECD network with margin $m$, which is employed in the DLL for training the ADCMH parameters in stage 1 (a). In stage 2 of the alternative minimization algorithm, intermediate hash codes generated by the ADCMH network (i.e., from stage 1(a)) are passed through the  trained NECD network (i.e., from stage 1(b))  to find the closest correct codeword to the intermediate hash codes. The cross-entropy loss between the correct codeword and the intermediate codes is then back-propagated only to the  ADCMH network to update its parameters. It should be noted that during the testing, only the ADCMH component of the DNDCMH is used for image retrieval and not the NECD component.

Specifically, We train neural error correcting decoder (NECD), which is a neural implementation of ECC, such that it has an error correcting capability greater than or equal to the margin in the  DCMH loss function and then use it to correct the intermediate hash codes generated by DCMH up to the Hamming distance of the margin. This allows us to improve the retrieval efficiency by correcting the hashing codes in the following ways: 1) If the intermediate image and attribute hash codes obtained at the output of the ADCMH belong to the same subject, the Hamming distance between  those intermediate hash codes will be less than the margin, which means that the distance between them is within the error correcting capability of the NECD. In this case, NECD will force and push those intermediate hash codes to decode to the same codeword. This helps to improve the retrieval efficiency as attribute and image belonging to the same subject will be pushed towards each other leading to more true positives. 2) Similarly, if the intermediate image and attribute hash codes obtained at the output of ADCMH belong to different subjects, the Hamming distance between the hash codes  will be greater than the margin, which means that the distance between them will fall outside the error correcting capability of the NECD and they will be decoded to different codewords, which implies attribute and image belonging to different subject will be pushed away from each other leading to less false positives.

 To summarize, the main contributions of this paper include: 

\textbf{1: Attribute guided deep cross-modal hashing (ADCMH)}: We utilize deep cross-modal hashing based on a margin-based DLL for face image retrieval in response to a facial attribute query.
 
\textbf{2: Correcting cross-modal hashing codes using a neural error-correcting decoder (NECD)}: We exploit the error correcting capability of an ECC and relate it to the margin of DLL to integrate the NECD network into the ADCMH network to learn error-corrected hash codes using an alternative minimization optimization algorithm.
 
 \textbf{3: Scalable cross-modal hash}: The proposed DNDCMH architecture performs facial image retrieval using point-wise data without requiring pairs or triplets of training inputs, which makes DNDCMH scalable to large scale datasets.
 
 \section{Related Work}
 
 \subsection{Retrieving Facial Images for an  Attribute Query}

Searching for facial images in response to a facial attribute query has been investigated in the past \cite{kumar2008facetracer, 2011_siddiquie_image_ranking,2009_vaquero_attribute}. FaceTracer, an image search engine that allows users to retrieve face images  based on queries involving multiple visual attributes was built in \cite{kumar2008facetracer} using a combination of support vector machines and Adaboost. Owing to the challenges of face recognition in surveillance scenarios, Vaquero \etal \ \cite{2009_vaquero_attribute}  proposed to search for people in surveillance systems based on a parsing of human parts and their attributes, including facial hair, eyeglasses, clothing color, etc.  However, in \cite{kumar2008facetracer}, and \cite{2009_vaquero_attribute}, the correlation between attributes is not considered. To overcome this problem, Siddiquie \etal \ \cite{2011_siddiquie_image_ranking} proposed a ranking and image retrieval system for faces based on multi-attribute queries, which explicitly modeled the correlations that are present between the attributes. 
 \begin{figure*}[t]
\centering
\includegraphics[width=18cm]{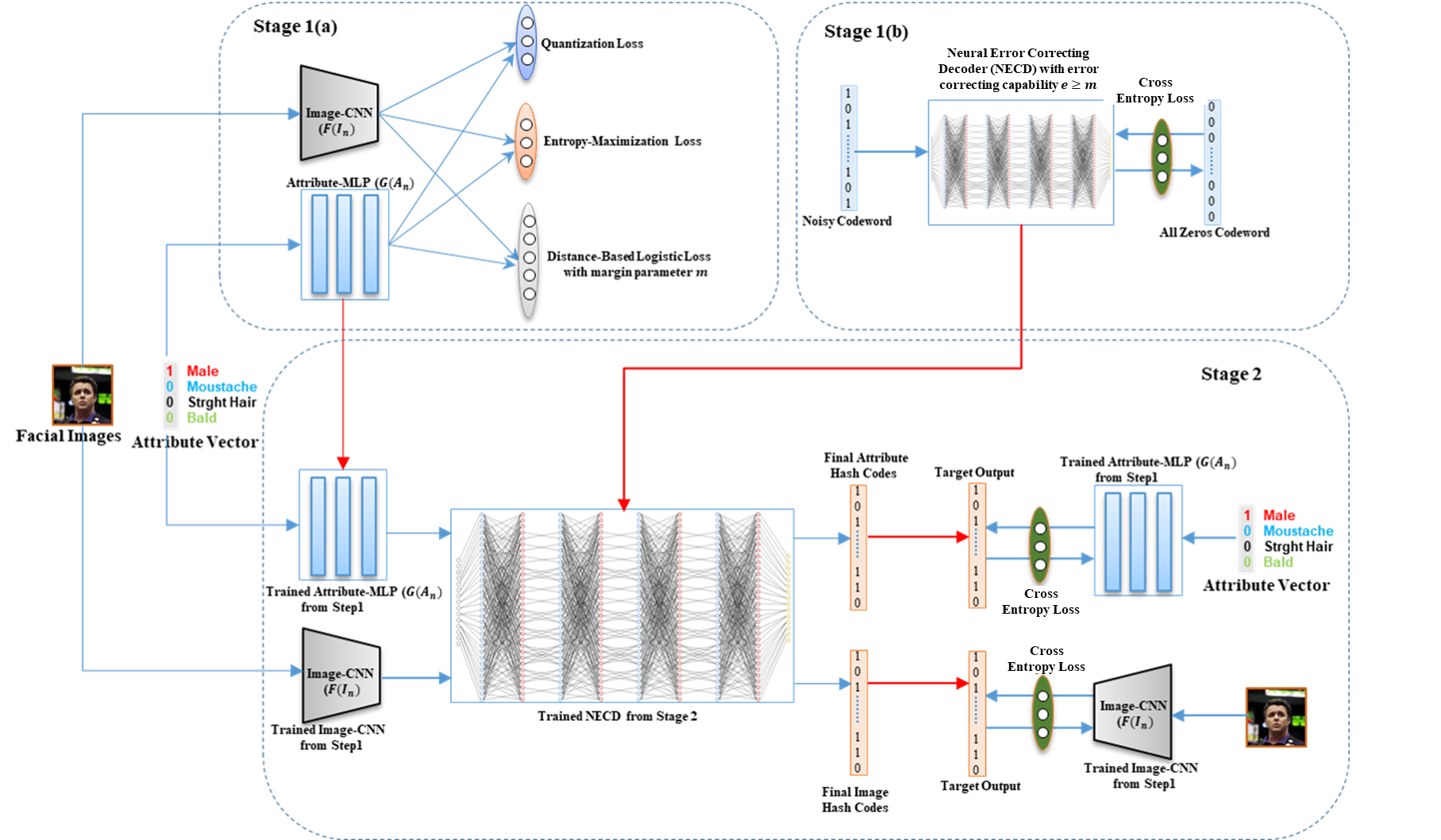}
\caption{Schematic illustration of our DNDCMH. It consists of two networks, the ADCMH and NECD. The ADCMH network consists of the Image CNN and the Attribute-MLP. The training of the DNDCMH is performed in 2 stages. The first stage is divided into: Stage 1(a) and Stage 1(b). In Stage 1(a), the Attribute-MLP and Image CNN of the ADCMH network are trained together with the quantization loss, entropy maximization and distance-based logistic loss with margin $m$. In Stage 1(b), NECD with an error-correcting capability of $e \geq m$ is trained using the cross-entropy loss. In Stage 2, the trained ADCMH from Stage 1(a) and trained NECD from Stage 1(b) are used (this is indicated by the red arrows). In Stage 2, the same training data as used in Stage 1(a) is passed through the trained ADCMH and the real-valued output of the ADCMH is then passed through the trained NECD to provide the final hash codes. Next, these final hash codes are used as target outputs (final hash codes and target outputs are the same as indicated by the red arrows) to optimize the trained ADCMH network. The optimization of the ADCMH network in this Stage is performed by using cross-entropy loss. The blue arrows indicates the inputs and the connections between the network and the losses. This is only the training process, the testing process is shown in Fig. \ref{fig:testing}
}\label{fig:new_arch}
\vspace{-0.40cm}
\end{figure*}
 \subsection{Cross-Modal Hashing}
Cross-modal hashing (CMH) can generally be divided into two categories: unsupervised hashing and supervised hashing. Unsupervised hashing \cite{2014_ding_cmfh, 2017_liu_cross_modality,lin2018unsupervised,Su_2019_ICCV,8734835,li2020semi} relates to learning hash codes from unlabeled data, while attempting to preserve semantic similarity between the data samples in the original space. In collective matrix factorization hashing (CMFH) \cite{2014_ding_cmfh}, collective matrix factorization is used to learn unified hash codes in a latent Hamming space shared by both modalities. In fusion similarity hashing (FSH) \cite{2017_liu_cross_modality}, an undirected asymmetric graph is leveraged to model the fusion similarity among different modalities.
 
 On the other hand, supervised hashing methods \cite{2010_bronstein_cmssh,Zhang_2014_LSM,Zhen_NIPS2012_4793,xu2019graph,Li_2018_CVPR}  take full advantage of the label information to mitigate the semantic gap and improve the hashing quality, therefore attaining higher search accuracy than the unsupervised methods. In semantic correlation maximization hashing (SCMH) \cite{Zhang_2014_LSM}, semantic labels are merged into the hash learning procedure for large-scale data modeling. In co-regularized hashing (CRH) \cite{Zhen_NIPS2012_4793}, each bit of the hash codes are learned by solving the difference of convex functions programs, while the learning for multiple bits is performed by a boosting procedure.

In recent years, deep learning methods have shown impressive learning ability in image recognition \cite{he_resnet_2016,krizhevsky_imagenet_2012,simonyan_very_deep_2014,szegedy_googlenet_2014}, object detection \cite{Erhan_2014_CVPR,ren_faster_rcnn_2015}, speech recognition \cite{hinton_DNN_speech_2012,graves_sppech_recog_2013} and many other computer vision tasks. The application of deep learning to hashing methods improves performance. There also exist methods (eg., \cite{Yang_2017_Pairwise,jiang2017deep}) which adopt deep learning for cross-modal hashing (CMH) and give improved performance over other CMH techniques that use handcrafted features \cite{Zhang_2014_LSM,Zhen_NIPS2012_4793}. Jiang \etal \ were the first to propose an end-to-end deep cross-modal hashing framework to learn the binary hash codes in DCMH \cite{jiang2017deep}. However, they just utilize the inter-modal relationship ignoring intra-modal information. In contrast, Yang \etal \ exploit this intra-modal information by using pairwise labels to propose Pairwise Relationship Guided Deep Hashing (PRDH) \cite{Yang_2017_Pairwise}.

\subsection{Neural Error-Correcting Decoder}
In addition to DCMH, the other deep learning network that our system uses is neural error-correcting decoder (NECD). In \cite{2016_nachmani_NND}, the BP algorithm is formulated as a neural network and it is shown that a weighted BP decoder implemented by deep learning methods can improve the BP decoding of codes by 0.9 dB in the high signal to noise ratio (SNR) region. Later, Lugosch \etal \ \cite{Lugosch_2017_NeuralOM} proposed a neural network architecture with reduced complexity by leveraging the offset min-sum algorithm and achieved similar results to \cite{2016_nachmani_NND}. Gruber \etal \ \cite{Gruber_17_arxiv} used a fully connected architecture to propose a neural network decoder that gives performance close to a maximum likelihood (ML) decoder for very small block codes. Additionally, in \cite{Shea_2017_physical}, a communication system has been formulated as an autoencoder for a small block code.





\section{Proposed Method}

In this section, we first formulate the problem; then, we
provide the details of our proposed method including the cross-modal hashing framework, training of our proposed system, and how it can be leveraged for out-of-sample data.

\subsection{Problem Definition}\label{ssec:probdef}

\color{black} We assume that there are two
modalities for each sample, i.e., facial attribute and image. Define $\textbf{X} = \{\textbf{x}\textsubscript{i}\}_{i=1}^{n}$ to represent the image modality, in which \textbf{x}\textsubscript{i} is the raw pixels of image $i$ in a training set $X$ of size $n$. In addition, we use $\textbf{Y} = \{\textbf{y}\textsubscript{i}\}_{i=1}^{n}$ to represent the attribute modality, in which \textbf{y}\textsubscript{i} is  the annotated facial attributes vector related to image $\textbf{x}_i$. $\textbf{S}$ is a cross-modal similarity matrix  in which $S_{ij} = 1 $ if image $\textbf{x}\textsubscript{i}$ contains a facial attribute y\textsubscript{j}, and $S_{ij} = 0$ otherwise.

Based on the given training information (i.e., \textbf{X}, \textbf{Y} and \textbf{S}),  the  goal of our proposed method is to learn modality-specific hash functions: $h^{(x)} (\textbf{x}) \in \{-1, +1\} ^ c $ for image modality, and $h^{(y)} (\textbf{y}) \in \{-1, +1\} ^ c $ for attribute modality to map the image $\textbf{x}$ and attribute feature vector $\textbf{y}$ into a compact $c$-bit hash code. Hash codes need to be learned such that the cross-modal similarity in \textbf{S} is preserved in Hamming space. Specifically, if $S_{ij} = 1$, the Hamming distance
between the binary codes $ \textbf{c}^{(\textbf{x}_{i})}_{i}= h^{(x)} (\textbf{x}\textsubscript{i}) $ and $\textbf{c}^{(\textbf{y}_{j})}_{j}= h^{(y)} (\textbf{y}\textsubscript{j}) $ should be small and if $S_{ij} = 0$, the corresponding Hamming distance should be large.

\subsection{Deep Neural Decoder Cross-Modal Hashing}\label{subsec:DNDCMH}

A schematic of our proposed DNDCMH is shown in Fig. \ref{fig:new_arch}. The DNDCMH architecture consists of two important components: 1) An ``attribute-based deep cross-modal hashing" (ADCMH) architecture, which contains an image convolutional neural network (Image-CNN), and an attribute multi-layer perceptron (Attribute-MLP). 2) A neural error-correcting decoder (NECD). As shown in Fig. \ref{fig:new_arch}, the entire DNDCMH network is trained end to end  by implementing an alternative minimization algorithm in two stages. 

Stage 1 is split into parts 1(a) and 1(b) such that stage 1(a) of this algorithm, learns ADCMH parameters by using a novel cross-modal loss function that uses margin-based distance logistic loss (DLL) to learn the ADCMH parameters. Stage 1(a) of the algorithm generates intermediate cross-modal hash codes. Stage 1(b) learns a NECD by relating the error correcting capability $e$ of the ECC used to create the NECD with the margin $m$ used in the distance logistic loss for training the ADCMH parameters in stage 1(a). Specifically, in stage 1(b), in order to force the attribute and image intermediate hash codes of the same subject to be pushed closer and decoded to the same codeword, we choose the ECC associated with the NECD in such a way that the error correcting capability $e$ is greather than or equal to $m$, where $m$ is the margin parameter used in the distance logistic loss for training the ADCMH parameters in stage 1(a).

In stage 2 of alternative minimization algorithm, we pass the intermediate hash codes (real-valued vector) generated by ADCMH (from stage 1(a)) to the  trained NECD network (from stage 1(b))  to find the closest correct codeword (binary) to the intermediate hash codes. The cross-entropy loss between the correct codeword and the intermediate codes is back-propagated only to the  ADCMH network to update its parameters. Stage 1(a) and stage 2  are  done iteratively until there is no longer a significant improvement  in retrieval efficiency with respect to the training.

\subsubsection{Stage 1(a): Learning Intermediate Hash Codes}

\textbf{Training of ADCMH network to learn the intermediate hash codes}: In stage 1(a), the ADCMH network, which is a coupled deep neural network (DNN), is trained to learn the intermediate hash codes. The ADCMH has three main objectives: 1) to learn a coupled DNN using distance-based logistic loss to preserve the cross-modal similarity between different modalities; 2) to secure a high retrieval efficiency, for each modality, minimize the quantization error due to the hashing of real-valued continuous output activations of the network to binary codes; 3) to maximize the entropy corresponding to each bit to obtain the maximum information provided by the hash codes.

\color{black} We design the objective function for ADCMH to generate efficient hash codes. Our objective function for ADCMH  comprises of  three parts: (1) margin-based distance logistic loss; (2) quantization loss; and (3) entropy maximization loss. The ADCMH is composed of two networks: An Image-CNN, which is used to extract features for image modality and an Attribute-MLP, which is used to extract features for facial attribute modality. A $\mbox{tanh}$ activation is used for the last layer of both networks so that the network outputs are in the range of [-1,1].  Let $ p(\textbf{w}_x, \textbf{x}\textsubscript{i}) \in \mathbb{R}^d $ denote the learned CNN features for sample $\textbf{x}\textsubscript{i}$  corresponding to the image modality, and  $ q(\textbf{w}_y, \textbf{y}\textsubscript{j}) $ denote the learned MLP features for sample $\textbf{y}\textsubscript{j}$ corresponding to the attribute modality. $\textbf{w}_x$ and $\textbf{w}_y$ are the CNN network weights and the MLP network weights, respectively.
The total objective function for ADCMH is defined as follows:
\begin{equation}
\begin{split}
     \mathcal{J} & = \sum_{i=1}^n \sum_{j=1}^n   \ell_c( p(\textbf{P}_{*i},\textbf{Q}_{*j}),S_{ij}) \\ & - \frac{\theta}{c} ( \sum_{i=1}^n||\textbf{P}_{*i}||^2+\sum_{j=1}^n||\textbf{Q}_{*j}||^2) 
    \\ & + \lambda(\sum_{e=1}^c||(\textbf{P}^\mathsf{T})_{*e}||^2+\sum_{f=1}^c||(\textbf{Q}^\mathsf{T})_{*f}||^2) ,
\end{split} \label{eq:1}   
\end{equation}
where $\textbf{P} \in  \mathbb{R}^{c \times n}$ represents the image feature matrix constructed by placing  the CNN features of the training samples column-wise and $\textbf{P}_{*i}= p(\textbf{w}_x, \textbf{x}\textsubscript{i})$ is the CNN feature corresponding to sample $\textbf{x}\textsubscript{i}$. $(\textbf{P}^\mathsf{T})_{*e}$ is a column vector, representing the $e$-th bit of all the training samples. Likewise, $\textbf{Q} \in  \mathbb{R}^{c \times n}$ represents a facial attribute feature matrix and $\textbf{Q}_{*j}= q(\textbf{w}_y, \textbf{y}\textsubscript{j})$ is the MLP feature for the attribute modality $\textbf{y}\textsubscript{j}$; $(\textbf{Q}^\mathsf{T})_{*f}$ is a column vector, which represents the $f$-th bit of all the training samples. The objective function in (\ref{eq:1}) needs to be minimized with respect to parameters $\textbf{w}_x,\textbf{w}_y$.

 \begin{figure}[t]
\centering
\includegraphics[width=6cm,height=5cm]{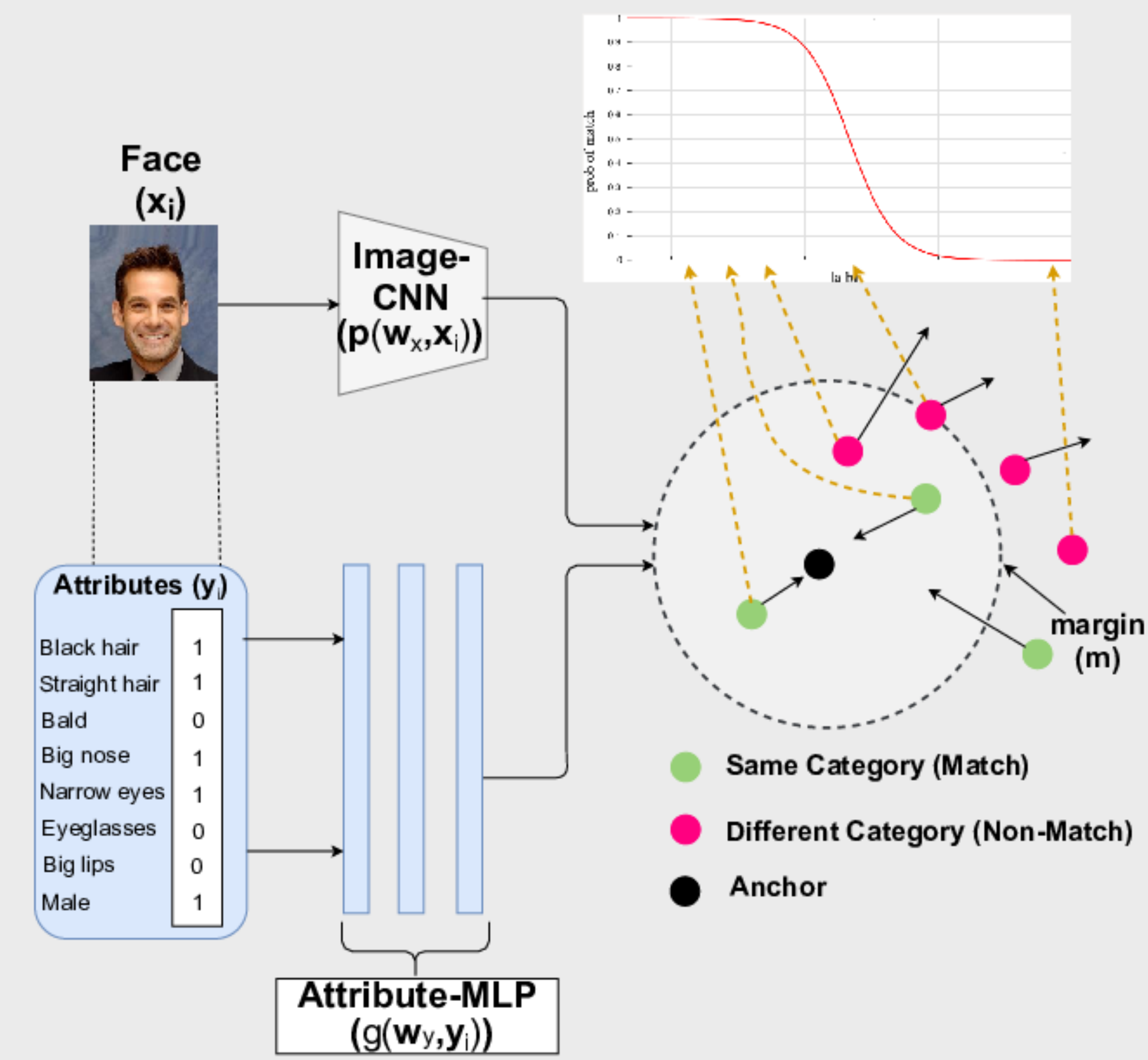}
\caption{Diagram for the first stage showing the importance of margin-based distance logistic loss. }\label{fig:stg_1}

\end{figure}

The first term in the objective function is the margin-based DLL, which tries to push modalities referring to the same sample closer to each other, while pushing the modalities referring to different samples away from each other. The term $p(\textbf{P}_{*i},\textbf{Q}_{*j})=\frac{1+\exp(-m)}{1+\exp(||\textbf{P}_{*i}-\textbf{Q}_{*j}||-m)}$ represents the distance-based logistic probability (DBLP) and defines the probability of the match between the image modality feature vector $\textbf{P}_{*i}$ and attribute modality feature vector $\textbf{Q}_{*j}$, given their squared distance. The margin $m$ determines the extent to which matched or non-matched samples are attracted or repelled from each other, respectively. The distance-based logistic loss is then derived from the DBLP by using the cross-entropy loss similar to the classification case: $\ell_c(r,s)=-s \log (r) +(s-1) \log(1-r)$.
  \begin{figure*}[t]
\centering
\subfigure[Error-Correcting capability]{\includegraphics[scale=0.40]{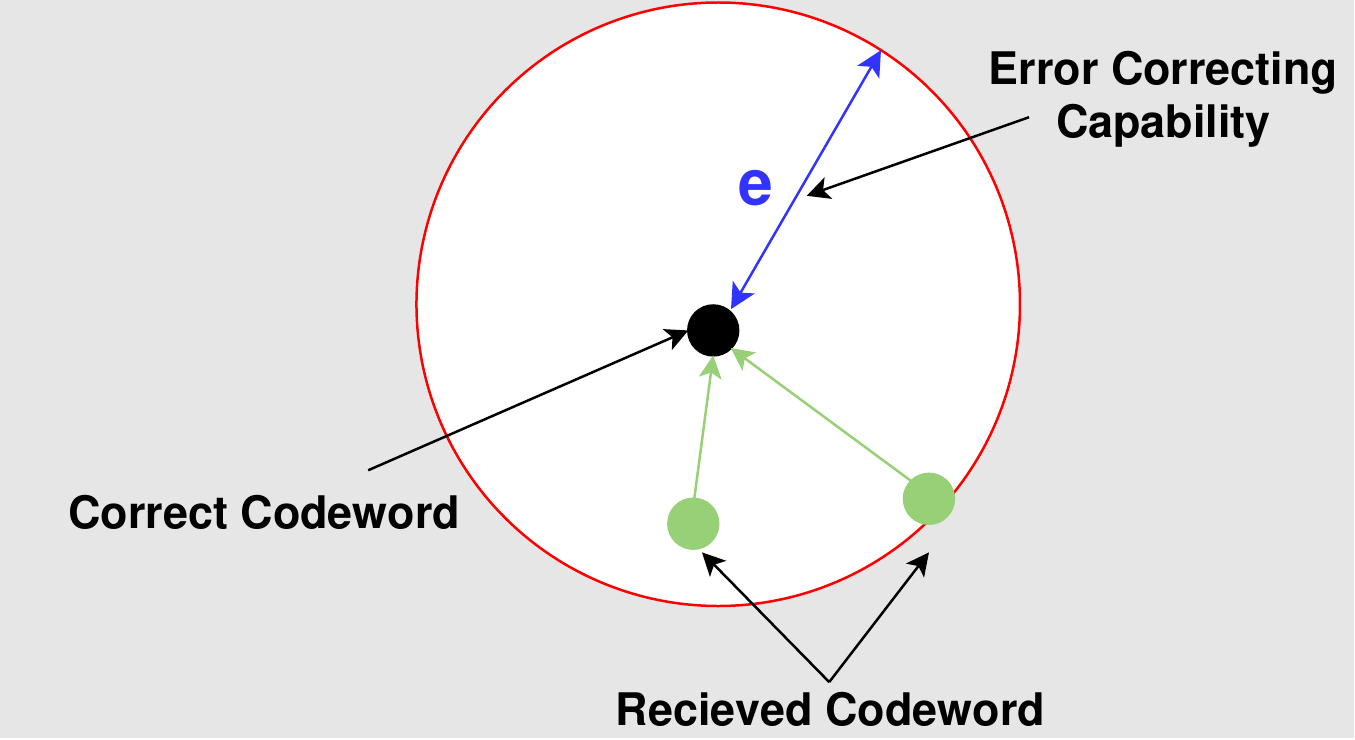}\label{fig:stg_2}}
\subfigure[Margin of DLL]{\includegraphics[scale=0.455]{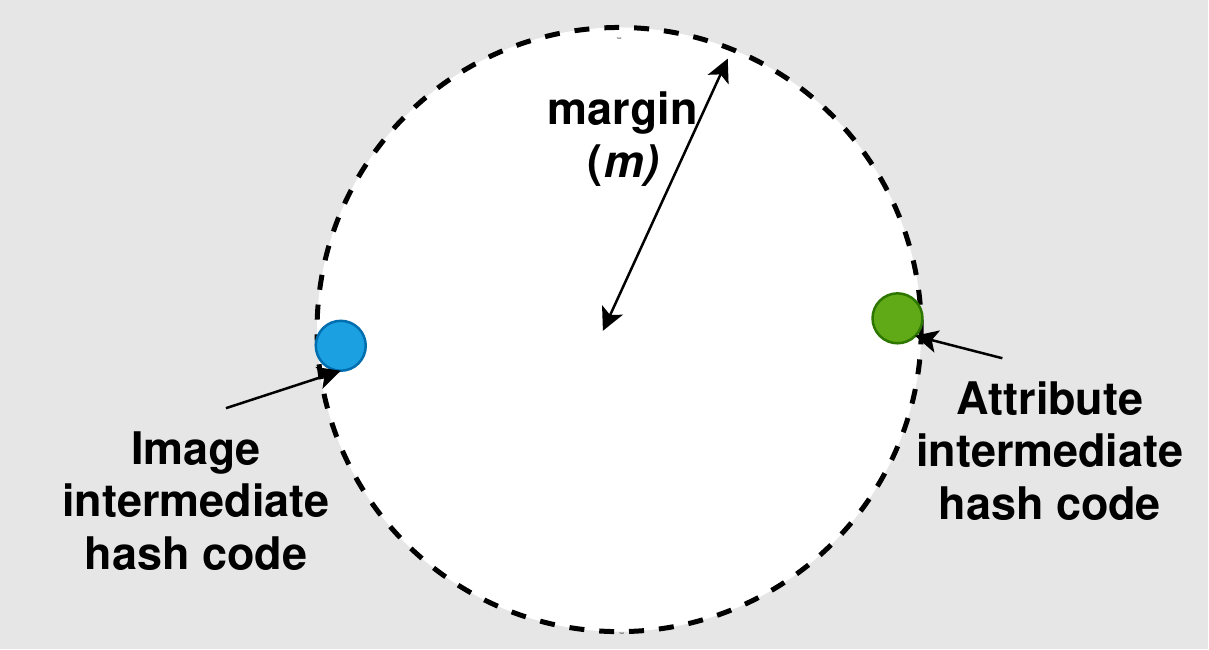}\label{fig:stage_1(b)_1}}
\subfigure[Margin of DLL and the error-correcting capability ]{\includegraphics[scale=0.443]{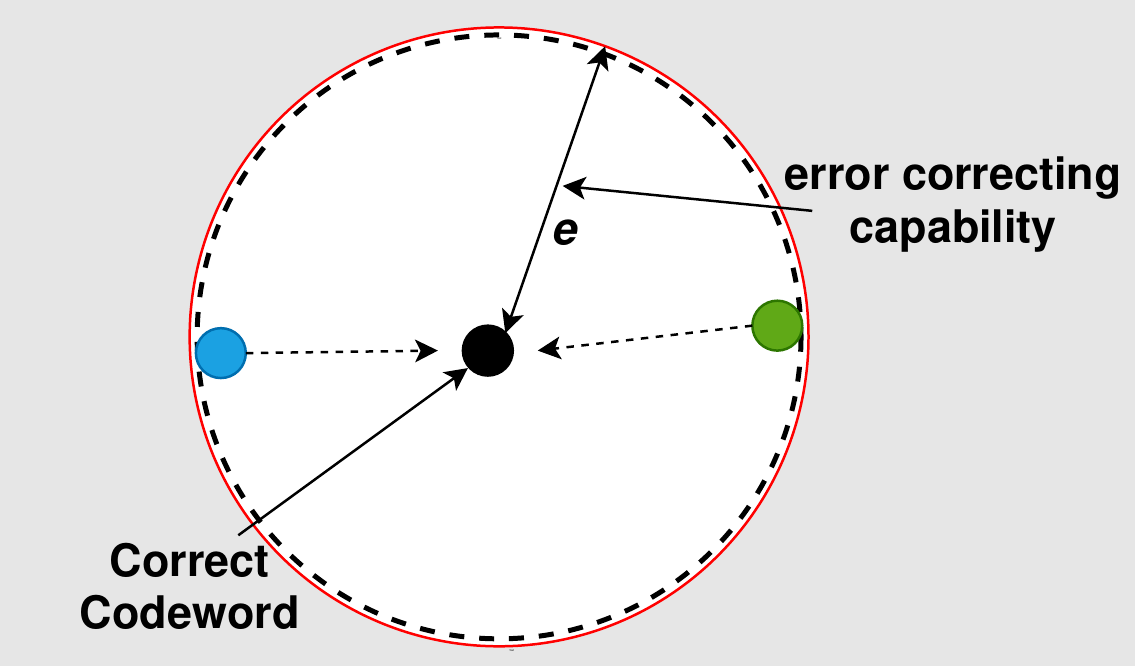}\label{fig:stage_1(b)_2}}

\caption{Relating the error-correcting capability of ECC to the margin of DLL.}

\label{fig:stage_1(b)}
\end{figure*}
Fig. \ref{fig:stg_1} shows an illustration for the margin-based distance logistic-loss for our application. The dotted circle indicates the margin $m$ to the boundary in terms of Hamming distance. The anchor (black solid circle) indicates a fixed instance and could either be an image hash code or an attribute hash code. The green solid circles indicate the matched instances (i.e., image or attribute hash codes belonging to the same subject as the anchor) and the magenta solid circles indicate non-matched instances (i.e., image or attribute hash codes belonging to different subject than the anchor). The function of margin-based DLL is two-fold. First, it pushes the matched instances (green circles) away from the margin in the inward direction i.e., the Hamming distance between the image and attribute hash codes of the same subject should be less than the margin $m$. Second, the margin-based DLL also pushes the non-matched instances (magenta circles) away from the margin in the outward direction i.e., the Hamming distance between the image and attribute hash codes belonging to different subjects should be larger than the margin $m$. It has to be noted that after training the ADCMH network, the image hash codes and the attribute hash codes belonging to the same subject have a Hamming distance less than margin $m$. This margin is a very important parameter in this framework and it will also affect the training of  NECD as is detailed in the description of stage 1(b) in Sec. \ref{subsec:1(b)} .

The second term in the objective function is the quantization loss that incentivizes the activations of the units in the hashing layer to be closer to -1 or 1. The values of the elements of $\textbf{P}_{*i}$ are in the range of $[-1,1]$ because they have been activated by the $\mbox{tanh}$ activation. To make the codes closer to either -1 or 1, we add a quantization constraint of maximizing the sum of squared errors between the hashing layer activations and 0, which is given by $\sum_{i=1}^{n}||\textbf{P}_{*i}-\textbf{0}||^{2}$, where $n$ is the number of training images in a mini-batch and \textbf{0} is the $c$-dimensional vector with all elements equal to 0. However, this is equivalent to maximizing the square of the length of the vector formed by the hashing layer activations, that is $\sum_{i=1}^{n}||\textbf{P}_{*i}-\textbf{0}||^{2}=\sum_{i=1}^{n}||\textbf{P}_{*i}||^{2}$. 


The third term, the entropy maximization loss, helps to obtain hash codes with an equal number of -1’s and 1’s, which maximizes the entropy of the discrete distribution and results  in  hash  codes  with  better  discrimination. Precisely, the number of +1 and -1 for each bit on all the training samples should be almost the same.


\textbf{Learning the ADCMH parameters in Stage 1(a):}
We used an alternating minimization algorithm to learn the ADCMH network parameters $\textbf{w}_x$, and $\textbf{w}_y$. In this algorithm, within each epoch, we learn one parameter with other parameters fixed. 


\textbf{Learning ($\textbf{w}_x$) parameters for ADCMH} : We use the back propagation algorithm to first optimize the CNN parameters $\textbf{w}_x$ for the image modality by fixing the $\textbf{w}_y$, and compute loss function gradient with respect to the output of image modality network as follows:
\begin{equation}
\begin{split}
    \frac{\partial \mathcal{J}}{\partial \textbf{P}_{*i}}=\sum_{j=1}^n \frac{\partial \ell_c(p(\textbf{P}_{*i}, \textbf{Q}_{*j}), S_{ij})}{\partial \textbf{P}_{*i}} & - \frac{2\theta}{c} \sum_{i=1}^n(\textbf{P}_{*i})\\ &+2\lambda\sum_{i=1}^c(\textbf{P}^\mathsf{T})_{*i}  \end{split} \label{eq:2}.
\end{equation} The gradient of the first term in Eq. \ref{eq:2} is calculated as:
\begin{equation}
\begin{split}
   \frac  {\partial \ell_c( p(\textbf{P}_{*i},\textbf{Q}_{*j}),S_{ij})}{\partial \textbf{P}_{*i}} &= \frac{-(1+\exp(-m))}{(1+\exp(||\textbf{P}_{*i}-\textbf{Q}_{*j}||-m))^2} \\ & \times (\frac{S_{ij}}{p(\textbf{P}_{*i},\textbf{Q}_{*j})}+\frac{1-S_{ij}}{1-p(\textbf{P}_{*i},\textbf{Q}_{*j})})
   \end{split}
   .  
\end{equation}
Next, we  compute $\frac{\partial \mathcal{J}}{\partial \textbf{w}_x} $ with $ \frac{\partial \mathcal{J}}{\partial \textbf{P}_{*i}}$ by using the chain rule ($\frac{\partial \mathcal{J}}{\partial \textbf{w}_{x}}= \frac{\partial \mathcal{J}}{\partial \textbf{P}_{*i}} \times  \frac{\partial \textbf{P}_{*i}}{\partial \textbf{w}_{x}} $), based on which the back propagation is  used to update the parameter $\textbf{w}_x$.

\textbf{Learning ($\textbf{w}_y$) parameters for ADCMH} : Similar to the previous optimization, we use the back propagation algorithm  to optimize  the MLP network parameters $\textbf{w}_y$ for the facial attribute modality by fixing  $\textbf{w}_x$, and compute the loss function gradient with respect to the output of the facial attribute network as follows:
\begin{equation}
\begin{split}
    \frac{\partial \mathcal{J}}{\partial \textbf{Q}_{*j}}=\sum_{i=1}^n \frac{\partial \ell_c(p(\textbf{P}_{*i}, \textbf{Q}_{*j}), S_{ij})}{\partial \textbf{Q}_{*j}} & - \frac{2\theta}{c} \sum_{j=1}^n(\textbf{Q}_{*j})\\ &+2\lambda\sum_{j=1}^c(\textbf{Q}^\mathsf{T})_{*j} \end{split} \label{eq:3}.
\end{equation}

Next, we  compute $\frac{\partial \mathcal{J}}{\partial \textbf{w}_y} $ with $ \frac{\partial \mathcal{J}}{\partial \textbf{Q}_{*j}}$ by using the chain rule ($\frac{\partial \mathcal{J}}{\partial \textbf{w}_{y}}= \frac{\partial \mathcal{J}}{\partial \textbf{Q}_{*j}} \times  \frac{\partial \textbf{Q}_{*j}}{\partial \textbf{w}_{y}} $), based on which the back propagation algorithm is  used to update the parameters $\textbf{w}_y$. 



The intermediate hash codes after training of the ADCMH are obtained by $\textbf{c}_{i}^{(\textbf{x}_{i})}=\mbox{sign}(p(\textbf{w}_{x},\textbf{x}_i))$ and $\textbf{c}_{i}^{(\textbf{y}_{i})}=\mbox{sign}(q(\textbf{w}_{y},\textbf{y}_i))$ for image $\textbf{x}_i$ and attribute $\textbf{y}_i$, respectively for a new sample outside of the training set.

\subsubsection{Stage 1(b): Training Neural Error Correcting Decoder} \label{subsec:1(b)}


  As mentioned previously, there is room to improve the cross-modal retrieval efficiency by reducing the Hamming distance between the intermediate hash codes of different modalities for the same subject. This can be achieved by using an ECC decoder. The concept of ECC decoder is illustrated in Fig. \ref{fig:stg_2}. It is observed from the figure that if the received codewords or the corrupted codewords (green solid circles) fall within the error correcting capability $e$ of the ECC, then the received codeword is decoded to the correct codeword (black solid circle) by the ECC decoder.

  The image or the attribute intermediate hash codes generated by ADCMH can be considered to be corrupted codewords within a certain distance $d$ of a correct codeword of an ECC. If this distance $d$ is within the error-correcting capability $e$ of the ECC, then the ECC decoder will decode the intermediate hash codes  to corresponding correct codeword of the ECC. However, decoding the intermediate hash codes to a correct codeword does not assure an improvement in cross-modal retrieval efficiency. For improving the retrieval efficiency, in addition to the intermediate hash codes being decoded to the correct codeword of the ECC, we also require facial and attribute intermediate hash codes of the same subject to be decoded to the same codeword.
  
For fulfilling the above requirement, we exploit the error-correcting capability of the ECC decoder and relate it to the margin $m$ used in the DLL for training the ADCMH in stage 1(a). Consider  Fig. \ref{fig:stage_1(b)_1} which shows the circle representing the margin for the DLL loss in stage 1(a). The blue and green circles on the margin represent the image intermediate hash code and attribute intermediate hash code, respectively of the same subject. Now consider  Fig. \ref{fig:stage_1(b)_2}, which shows the margin circle (black dashed line) along with the Hamming sphere of the ECC (red solid line) with error correcting capability of $e$. We notice that the blue and green circle (i.e., image intermediate hash code and attribute intermediate hash code)  fall within the error correcting capability of the ECC. Consequently, the image and attribute intermediate hash codes will be decoded to the same correct codeword (i.e. black small circle in the center) by the ECC decoder. This scenario is feasible only when error correcting capability $e$ of the ECC decoder is at-least equal to the margin $m$ used for DLL. Therefore, we chose an ECC decoder in such a way that the error correcting capability of the ECC $e \geq m$.

 Recently, some excellent neural network based ECC decoders have been proposed \cite{2016_nachmani_NND,nachmani2017rnn,Lugosch_2017_NeuralOM},  which have achieved close to the maximum likelihood (ML) performance for an ECC. These methods can be leveraged to generate high-quality and efficient hash codes. Thus, we can adapt such a neural network based ECC decoder, train it, and use it as an ancillary component to refine the intermediate hash codes generated by ADCMH. For more details about ECC and BP decoding algorithm refer to Appendix \ref{app:a}. 
 







 \textbf{Neural error-correcting decoder}: The NECD is a non fully-connected neural network and can be considered as a trellis representation of the BP decoder in which the nodes of the hidden layer correspond to the edges in the Tanner graph. Let $N$ be the size of the codeword (i.e., the number of variable nodes in the Tanner graph) and $E$ be the number of the edges in the Tanner graph. This implies that the input layer of our NECD decoder consists of $N$ nodes. The input to each of the $N$ nodes is the channel log-likelihood ratio (LLR) corresponding to that particular variable node in the Tanner graph. All the hidden layers of the NECD have size $E$. A node in each hidden layer is associated with the message transmitted over some edge in the Tanner graph. The output layer of the NECD contains $N$ nodes that output the final decoded codeword.

The number of hidden layers in the NECD depends upon the number of iterations considered for the BP algorithm. One iteration corresponds to message passing from the variable node to the check node and again back from check node to the variable node. Let us consider $L$ iterations of the BP decoder. Then the number of hidden layers in the NECD would be equal to $2L$. Consider the $i$-th hidden layer, where $i=1,2,\cdots,2L$. For odd (even, respectively) values of $i$, each node in this hidden layer of the NECD corresponds to an edge $e=(v,c)$ connecting the variable node $v$ (check node $c$, respectively)  to the check node $c$ (variable node $v$, respectively) and the output of this node represents the message transmitted by the BP decoder over that edge in the Tanner graph. 

Next, we will discuss the way the NECD node connections are formed. A node in the first hidden layer (i.e., $i=1$) corresponding to the edge $e=(v,c)$ is connected to a single node in the input layer, which is the variable node $v$ associated with that edge. 
A processing node in the hidden layer $i$, where $i>1$ and $i$ is odd (even, respectively), corresponds to the edge $e=(v,c)$, and is connected to all the nodes in the layer $i-1$ associated with the edges $e'=(v,c')$ for $c'\neq c$ ($e'=(v',c)$ for $v'\neq v$, respectively). For an odd node $i$, a processing node in layer $i$, corresponding to the edge $e=(v,c)$ is also connected to the $v$th input node. We denote by $x_{i,e}$, the output of a processing node in the hidden layer $i$. In terms of the BP decoder, for an odd (even, respectively) $i$, this is the message produced after $\floor*{(i-1)/2}$ iterations, from variable to check (check to variable, respectively) node. 
For odd $i$ and $e=(v,c)$, we can use
\begin{equation}
    x_{i,e=(v,c)}=\mbox{tanh}\left(\frac{1}{2}\left(\textbf{w}_{i,v}l_{v}+\sum_{e'=(v,c'),c'\neq c}\textbf{w}_{i,e,e'}x_{i-1,e'}\right)\right),\tag{5}\label{eq:5}
\end{equation} under the initialization, $x_{0,e'}=0$ for all $e'$ (there is no information in the parity check nodes in the beginning). The summation in (\ref{eq:5}) is over all the edges $e'=(v,c')$ with variable node $v$ except for the target edge $e=(v,c)$. $\textbf{w}$ corresponds to the weight parameters of the neural network.
Similarly, for even $i$ and $e=(v,c)$, we use
\begin{equation}
    x_{i,e=(v,c)}=2\mbox{tanh}^{-1}\left(\prod_{e'=(v',c),v'\neq v}x_{i-1,e'}\right).\tag{6}\label{eq:6}
\end{equation} The final $v$th output of the network is given by

\begin{equation}
   o_{v}=\sigma\left(\textbf{w}_{2L+1,v}l_{v}+\sum_{e'=(v,c')}\textbf{w}_{2L+1,v,e'}x_{2L,e'}\right),\tag{7}\label{eq:7}
\end{equation} where $\sigma(x)=(1+e^{-x})^{-1}$ is a sigmoid function and is added so that the final network output is in the range $[0,1]$.

In stage 1(b), we build and train the NECD to be useful for correcting the intermediate hash codes generated by ADCMH in stage 1(a). In this regard, we select the ECC code to build the NECD is such a way that the error correcting capability $e$ of the ECC is at-least equal to the margin $m$. Based on this condition, we then train our NECD using a dataset of zeros codeword.

\subsubsection{Stage 2: Correcting the Intermediate Hash Codes}

As shown in Fig. \ref{fig:new_arch}, the trained ADCMH network and the trained NECD from stage 1(a) and stage 1(b), respectively, are utilized in stage 2 as shown by the red unidirectional arrows. In stage 2, we use the same dataset used in stage 1(a) and generate the real-valued vector output of the trained Image-CNN and trained Attribute-MLP (from stage 1(a)) without thresholding using sign function. This output from both the networks is then passed through the trained NECD from stage 1(b) to generate the corresponding image and attribute final binary hash codes. Next, the corresponding image and attribute final hash codes are used as target outputs (i.e., ground truths) to fine-tune the Image-CNN and Attribute-MLP, respectively from stage 1(a). As shown in Fig. \ref{fig:new_arch}, for fine-tuning the Image-CNN and Attribute-MLP, we use the intermediate hash codes (real-valued vector output of the trained Image-CNN and trained Attribute-MLP)  as our predicted outputs and use the corresponding final hash codes (binary) as our ground truths. We use cross-entropy as loss function to fine-tune the Image-CNN and Attribute-MLP in stage 2.  

\begin{equation}
  L_{C}(y,p)= -\frac{1}{N}\sum_{i=1}^N \sum_{j=1}^k y_{j}^{(i)} \log ( p_{j}^{(i)})+(1-y_{j}^{(i)}) \log (1-p_{j}^{(i)}),
\end{equation} where $y_{j}^{(i)}$ is the final hash code value for the $i$-${th}$ training sample and $j$-${th}$ element in the output layer. Similarly, $p_{j}^{(i)}$ is the intermediate hash code value for the $i$-${th}$ training sample and $j$-${th}$ element in the output layer. $N$ signifies the number of training samples in a mini-batch, $k$ signifies the hash code length, and $L_{C}(y,p)$ defines the cross-entropy loss function between the intermediate hash code vector $p$ and final hash code vector $y$.

In this stage of the training algorithm, the cross-entropy loss between intermediate hash codes generated by ADCMH and the corrected codewords obtained by NECD, is back-propagated to  the ADCMH network to update its parameters (i.e., $\textbf{w}_x$ and $\textbf{w}_y$). As mentioned earlier, intermediate hash codes used in this stage are real valued for both modalities. Here, we formulate stage 2 of the training algorithm, which is used to update the ADCMH parameters to  generate the corrected codes.

\begin{figure*}[h]
\centering
\includegraphics[width=9cm]{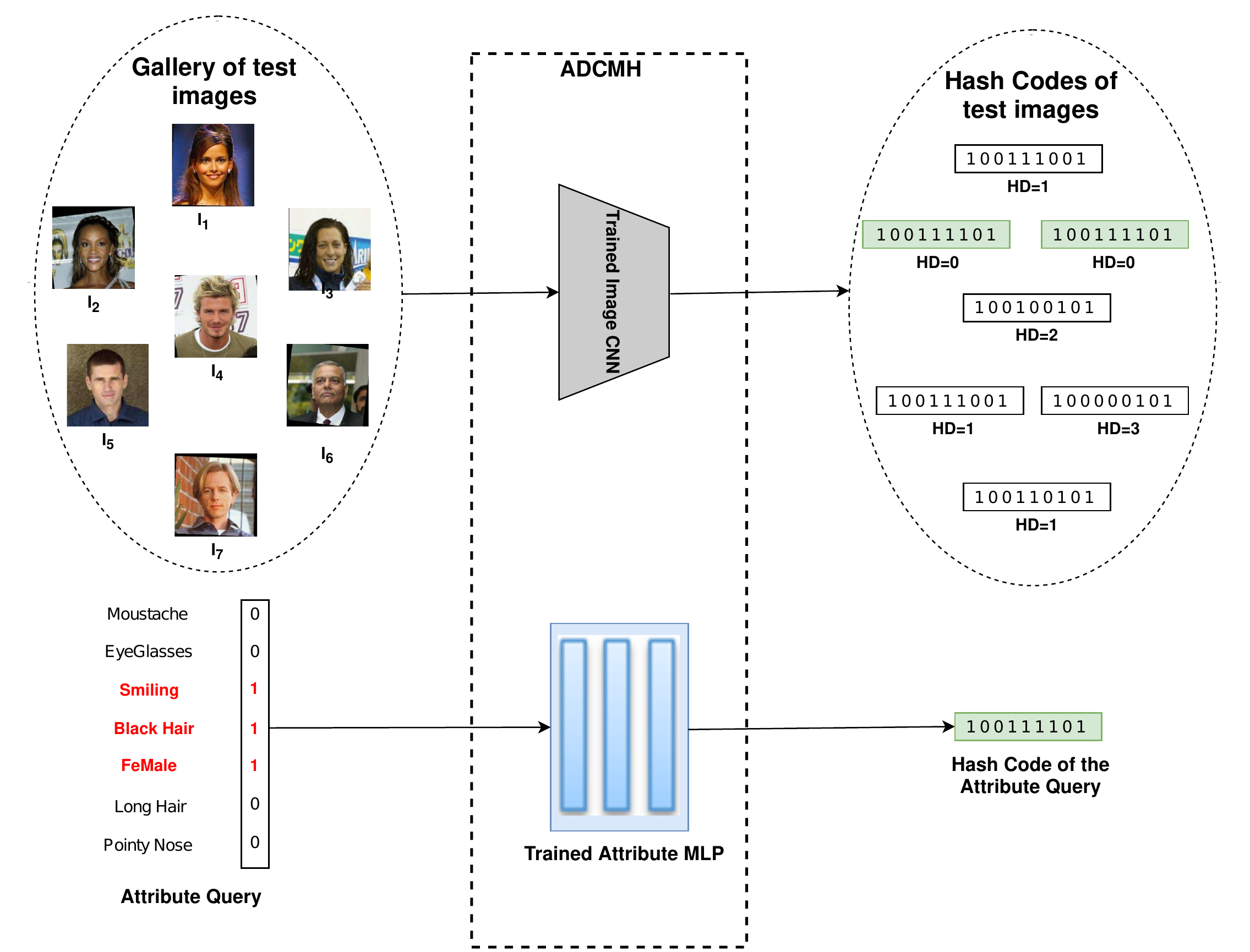}
\caption{Illustration of the testing process.  The gallery of test images is passed through the Image-CNN of the ADCMH network to generate the image hash code. The attribute query binary vector is passed through the trained Attribute-MLP to generate the attribute hash codes. The Hamming distance (HD) between the attribute hash code and the image hash code is computed. The HD between the each of the gallery image hash code and the attribute hash code is shown under the corresponding image hash code. The green color represents the lowest HD image hash code. The red color in the attribute query indicates the three attributes that we are interested in. It is a triple attribute query example. }\label{fig:testing}
\end{figure*}

As shown in Fig. \ref{fig:new_arch}, the NECD network is used to correct the intermediate codes generated from both modalities. Assume that $\textbf{t}_{i}^{(\textbf{x}_{i})}$ is the output of the NECD network used to correct the intermediate hash code $\textbf{c}_{i}^{(\textbf{x}_{i})}$ associated with the sample $\textbf{x}_i$ for image modality. The overall cross-entropy loss on  all the samples is calculated as: $ \mathcal{H}_p=L_C(\textbf{t}_{i}^{(\textbf{x}_{i})},\textbf{c}_{i}^{(\textbf{x}_{i})})$. To back-propagate the loss of the error-corrected code to the ADCMH parameters related to the image modality (i.e., $\textbf{w}_x$), we can use the chain rule as discussed in the previous section to update $\textbf{w}_x$, i.e., ($\frac{\partial \mathcal{H}_p}{\partial \textbf{w}_{x}}= \frac{\partial \mathcal{H}_p}{\partial \textbf{P}_{*i}} \times  \frac{\partial \textbf{P}_{*i}}{\partial \textbf{w}_{x}} $). 

Likewise, the NECD network is used to update the ADCMH network parameters that are related to the facial attribute modality (i.e., $\textbf{w}_y$). Therefore, we can formulate it as : $ \mathcal{H}_q=L_C(\textbf{t}_{i}^{(\textbf{y}_{i})},\textbf{c}_{i}^{(\textbf{y}_{i})})$. To back-propagate the loss of the error-corrected code to update $\textbf{w}_y$, we use the chain rule as  ($\frac{\partial \mathcal{H}_q}{\partial \textbf{w}_{y}}= \frac{\partial \mathcal{H}_q}{\partial \textbf{Q}_{*i}} \times  \frac{\partial \textbf{Q}_{*i}}{\partial \textbf{w}_{y}} $). Note that before back-propagating the cross-entropy loss to update the ADCMH parameters, we scale the obtained loss by a hyper-parameter $\gamma$ to create a balance between cross-entropy loss and the cross-modal hashing loss (i.e., $\mathcal{J}$ in (\ref{eq:1})) that is defined in step 1 of the training algorithm.

Note that we utilize alternative optimization algorithm and perform the stage 1(a) and stage 2 of the training algorithm iteratively  until there is not a significant
improvement in retrieval efficiency on the training set. Additionally, note that we use the same trained NECD from stage 1(b) in stage 2 for both the Image-CNN and the Attribute-MLP. Therefore, if the NECD in stage 1(b) has been created using ECC with an error-correcting capability of $e \geq m$, where $m$ is the margin used for DLL in stage 1(a), then the final hash codes would be the same for both the image and attribute modality at the end of the training of the alternative optimization algorithm. The complete algorithm is given in Fig. \ref{fig:new_arch}.

\subsubsection{Out-of-Sample Extension}
After training DNDCMH to convergence (i.e., no improvement in accuracy on the training set), for a new instance that is not in the training set, we can easily generate its error-corrected hash code as long as we can get one of its modalities. Given a query data point with image modality $\textbf{x}_i$, we directly use it as the input of the Image-CNN part of the trained DNDCMH, then forward propagate the query through the network to generate its hash code as: $\textbf{t}_{i}^{(\textbf{x}_{i})}=\mbox{sign}(p(\textbf{w}_{x},\textbf{x}_i)).$
Similarly, for the Attribute-MLP, we generate the
hash code of a data point with only attribute modality $\textbf{y}_i$ as: $\textbf{t}_{i}^{(\textbf{y}_{i})}=\mbox{sign}(q(\textbf{w}_{y},\textbf{y}_i)).$

For example, assume that we are provided with an attribute query in the form of an binary vector specifying the presence of certain facial attributes as it is shown in Fig. \ref{fig:testing}. Along with the attribute query, we are also provided with a gallery of facial images. We need to find all the faces in the given gallery that contain the attributes given in the query. In this case, the attribute query is passed through the trained Attribute-MLP and the attribute hash code is generated. Similarly, each facial image in the given gallery is passed through the Image-CNN to generate a hash code for each image. All the facial images are then ranked in order of increasing Hamming distance between the image hash code and the attribute hash code.

\begin{table*}
\centering
\small

\caption{MAP comparison tabulating the effect of the error correcting capability $e$ for a given margin $m$ and hash code length of 63 bits.}
\scalebox{0.80}{\begin{tabular}{|c|c|c|c|c|c|c|c|c|c|c|c|}
 \hline
\multicolumn{1}{|c|}{Margin} &\multicolumn{1}{|c|}{Code - Word Size}&\multicolumn{1}{|c|}{Error Correcting Capability}   &\multicolumn{3}{|c|}{CelebA} &\multicolumn{3}{|c|}{LFW}&\multicolumn{3}{|c|}{YouTube}\\ [0.5ex] 

 \cline{4-12}
 $(m)$ & $(N,K)$ & $(e)$ & Single & Double  & Triple  & Single & Double  & Triple & Single & Double  & Triple  \\ \hline \hline
 \multirow{5}{*}{3} & (63,51)& 2 & 63.215 & 61.779 & 57.349 & 60.239 & 58.322 & 56.115 & 59.334 & 57.976 & 56.254\\ \cline{2-12}
 
 & (63,45)& 3 &71.831& \textbf{68.335} & 66.131 & \textbf{67.117} & \textbf{65.383} & \textbf{64.118} & \textbf{66.1} & \textbf{64.579} & 62.215 \\ \cline{2-12}
 & (63,39)& 4 &\textbf{72.10}& 68.11 & \textbf{66.63} & 65.135 & 64.446 & 63.988 & 64.653 & 64.091 & \textbf{63.561} \\ \cline{2-12}
 & (63,36)& 5 &66.543& 64.671 & 62.156 & 61.90 & 57.783 & 55.9 & 60.117 & 59.098 & 59.22 \\ \cline{2-12}
 & (63,30)& 6 &61.10& 58.831 & 54.665 & 56.35 & 53.139 & 50.952 & 54.871 & 53.813 & 53.178 \\ \hline \hline
 \multirow{5}{*}{5} & (63,39)& 4 &70.55& 68.116 & 66.312 & 67.515 & 65.213 & 63.111& 65.334 & 63.578 & 62.11\\ \cline{2-12}
 
 & (63,36)& 5 &76.311& \textbf{74.132} & \textbf{70.877} & \textbf{71.31} & \textbf{69.156} & 66.932& \textbf{70.23} & \textbf{68.102} & 66.009 \\ \cline{2-12}
 & (63,30)& 6 &\textbf{77.19}& 72.066 & 69.992 & 70.482 & 67.354 & \textbf{67.131} & 69.998 & 67.521 & \textbf{66.217}\\ \cline{2-12}
 & (63,24)& 7 &72.634& 71.135 & 68.820 & 67.111 & 65.090 & 63.865 & 64.43 & 62.491 & 59.089\\ \cline{2-12}
 & (63,18)& 10 &60.225& 58.113 & 56.622 & 58.35 & 56.751 & 54.322 & 54.982 & 53.19 & 51.551\\ \hline \hline
  \multirow{5}{*}{6} & (63,36)& 5 & 69.121 & 66.398 & 63.1 & 64.334 & 63.901 & 63.178 & 63.456 & 58.01 & 56.41\\ \cline{2-12}
 
 & (63,30)& 6 & \textbf{79.125}& 73.167 & \textbf{72.913} & \textbf{74.873} & \textbf{71.242} & \textbf{70.643} & 71.998 & \textbf{68.01} & \textbf{65.389}\\ \cline{2-12}
 & (63,24)& 7 &76.339& \textbf{73.842} & 70.190 & 72.1 & 70.908 & 67.231 & \textbf{72.13} & 66.54 & 64.1\\ \cline{2-12}
 & (63,18)& 10 &63.12& 62.781 & 60.613 & 61.349 & 60.444 & 57.609 & 60.78 & 55.782 & 53.567\\ \cline{2-12}
 & (63,16)& 11 &55.1& 56.789 & 53.334 & 52.18 & 50.981 & 50.4 & 53.986 & 50.12 & 47.56\\ \hline \hline
 \multirow{5}{*}{7} & (63,30)& 6 & 77.351& 75.568 & 72.2211 & 73.181 & 71.362 & 71.61 & 72.367 & 69.08 & 66.890 \\ \cline{2-12}
 
 & (63,24)& 7 &\textbf{83.131}& \textbf{79.354} & \textbf{77.116} & \textbf{77.335} & \textbf{74.952} & \textbf{71.118}  & \textbf{78.80} & \textbf{75.47} & \textbf{70.721}\\ \cline{2-12}
 & (63,18)& 10 &76.541& 73.396 & 70.751 & 69.117 & 66.354 & 62.182 & 71.389 & 69.903 & 66.56\\ \cline{2-12}
 & (63,16)& 11 &72.225& 69.181 & 66.192 & 67.111 & 65.332 & 63.981 & 67.908 & 64.095 & 62.60\\ \cline{2-12}
 & (63,10)& 13 &62.25& 60.181 & 58.333 & 58.05 & 56.351 & 54.119 & 59.834 & 55.596 & 52.9\\ \hline 
\end{tabular}}
\label{table:e_vs_m}
\end{table*}

\section{Experimental Results}
\textbf{Implementation}: 
As mentioned previously, our proposed ADCMH is composed of two networks: An image convolutional neural network (Image-CNN), which is used to extract features for the image modality and an attribute multi-layer perceptron (Attribute-MLP), which is used to extract features for the facial attribute modality. We initialize our Image-CNN parameters with a VGG-19 \cite{simonyan_very_deep_2014} network pre-trained on the ImageNet \cite{deng2009imagenet} dataset. The original VGG-19 consists of five convolutional layers ($conv1-conv5$) and three fully-connected layers ($fc6-fc8$). We discard the $fc8$ layer and replace the $fc7$ layer with a new $f_{ch}$ layer with $c$ hidden nodes, where $c$ is the required intermediate hash code length. For the convolutional layers and $fc6$ layer we have used ReLU activation, for the $f_{ch}$ layer, we have used $\mbox{tanh}$  activation. The size of the intermediate hash length is equal to the size of the codeword used in the NECD network. 

The Attribute-MLP  contains three fully connected layers to learn the features for the facial attribute modality. To perform feature learning from the attributes, we first represent the attributes of each training sample as a binary vector, which indicates the presence or absence of corresponding facial attribute. This binary vector serves as a facial attribute vector and is used as an input to the Attribute-MLP. The first and second layers in the MLP network contain 512 nodes with ReLU activation and the number of nodes in the last fully connected layer is equal to the intermediate hash code length $c$ with $\mbox{tanh}$  activation.  The weights of the MLP network are initialized by sampling randomly from $\mathcal{N}(0,0.01)$ except for the bias parameters that are initialized with zeros. We use the Adam optimizer \cite{kingma2014adam} with the default hyper-parameter values ($\epsilon = 10^{-3}$, $\lambda_1 = 0.9$, $\lambda_2 = 0.999$) to train all the parameters using alternative minimization approach. The batch size in all the experiments is fixed to 128. ADCMH is implemented in TensorFlow with Python API and all the experiments are conducted on two GeForce GTX TITAN X 12GB GPUs.

\color{black}To train this NECD, it is sufficient to use training database constructed using noisy versions of a single codeword \cite{2016_nachmani_NND}. For convenience, we use noisy versions of zero codeword and our training database for NECD contains different channel output realization when the zero codeword is transmitted. The goal is to train NECD to attain $N$ dimensional output word, which is as close as possible to the zero codeword. We have trained our NECD on several codes including BCH(31,21), BCH(63,45), and BCH(127,92). This implies that the intermediate and the target hash code length for our experiments is equal to 31, 63, and 127. Note that the exact codeword size $(N,k)$ depends on the error correcting capability $e$ required for the NECD. Again, the error correcting capability $e$ depends on the margin $m$ of the DLL function used to train the ADCMH, such that $e \geq m$.  \color{black}  

 \textbf{Datasets}: We evaluated our proposed DNDCMH framework on three face datasets including the Labeled Faces in the Wild (LFW) \cite{huang2008labeled}, Large-scale CelebFaces Attributes (CelebA) Dataset \cite{liu_2015_celebA} , and YouTube faces (YTF) dataset \cite{wolf2011face}.

 LFW is a notable face database of more than 13,000 images of faces, created for studying the problem of unconstrained face recognition. The faces are collected from the web, detected and centered by the Viola Jones face detector.  CelebA  is a large-scale face attribute and richly annotated dataset containing more than 200K celebrity images, each of which is annotated with  40 facial attributes. CelebA has about ten thousand identities with  twenty images per identity on average. For comparison purposes, we have been consistent with the train and test split of these datasets as given on the dataset webpage.
 
 YTF \cite{wolf2011face}  is a dataset based on the name list of LFW but created for video based face recognition. All the videos in YTF were downloaded from YouTube. YTF dataset contains 3,425 videos from 1,595 identities. Each video varies from 48 to 6,070 frames, with an average length as 181.3 frames. We extract 60,300 frames for evaluation. Particularly, we use 40,000 frames with 420 identities for training and the other remaining 20,300 images with 198 identities for testing. Since  YTF  dataset  has  not  been  annotated  by  facial  attributes, we use a well-known attribute prediction model Mixed Objective Optimization Network (MOON) \cite{rudd2016moon}  to annotate  YTF with the  facial  attributes  and then use it in our model.
 
 
 
 \textbf{Baselines}: We have compared the retrieval and ranking performance of our system with some of the other state-of-the-art face image retrieval and ranking approaches including MARR \cite{2011_siddiquie_image_ranking}, rankBoost \cite{freund_2003_efficient}, TagProp \cite{guillaumin_2009_tagprop}. For fair comparison, we have exploited the VGG-19 architecture pretrained on Imagenet dataset, which is the same as the initial CNN of the image modality in DNDCMH, to extract CNN features. All the above baselines are trained based on these CNN features. Additionally, we have also compared our algorithm with the state-of-the-art deep cross modal hashing algorithms DCMH \cite{jiang2017deep}, pairwise relationship guided deep hashing for cross-modal retrieval (PRDH) \cite{Yang_2017_Pairwise}, and triplet-based hashing network (THN) \cite{deng2018triplet}. We have also compared the DNDCMH framework with only the ADCMH network; i.e., training using only stage 1(a).

\textbf{Evaluation Protocols}: For hashing-based retrieval, Hamming ranking is a  widely used retrieval protocol and we will also evaluate our DNDCMH method and compare it with other baselines using this protocol. The Hamming ranking protocol ranks the points in the database (retrieval set) according to their Hamming distances to a given query point, in an increasing order. Mean average precision (MAP) is the widely used metric to measure the accuracy of the Hamming ranking protocol. 

Similarly, for ranking protocol we use normalized discounted cumulative gain (NDCG) to compare ranking performance of DNDCMH with other baselines. NDCG is a standard single-number measure of ranking quality that allows non-binary relevance judgments. It is defined as  $ \mbox{NDCG}@ k=\frac{1}{Z} \sum_{i=1}^k\frac{2^{\mathsf{rel}(i)}-1}{\log(i+1)} $, where $\mathsf{rel}(i)$ is the relevance of the $i^{th}$ ranked image and Z is a normalization constant to ensure that the correct ranking results in an NDCG score of 1.
\begin{figure*}[t]
\centering
\subfigure{\includegraphics[scale=0.48]{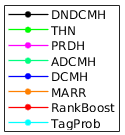}}
\subfigure[Single]{\includegraphics[scale=0.48]{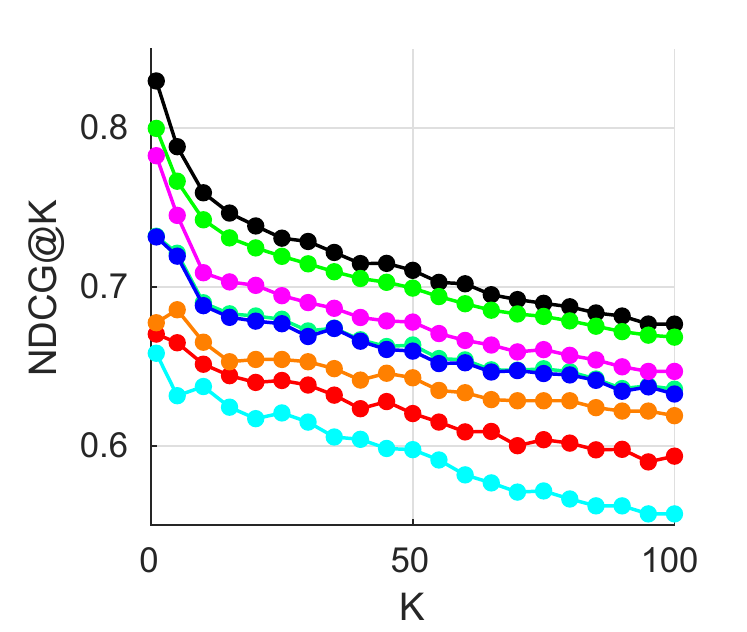}}
\subfigure[Double]{\includegraphics[scale=0.48]{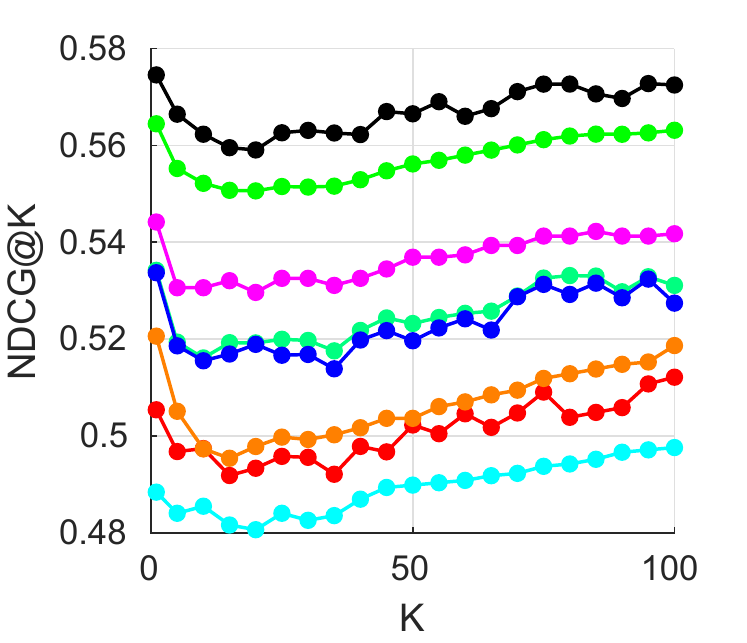}}
\subfigure[Triple]{\includegraphics[scale=0.48]{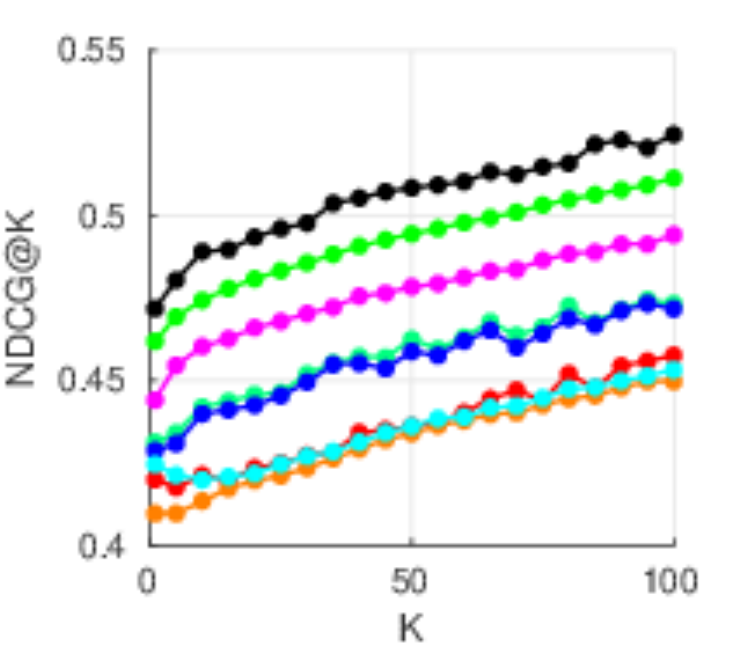}}
\caption{Ranking performance on the CelebA dataset. Due to space restriction, the legend is shown in the box on the left.}
\label{fig:Rank_LFW}
\vspace{-0.40cm}

\end{figure*}
\begin{figure*}
\centering
\subfigure{\includegraphics[scale=0.48]{legend.png}}
\subfigure[Single ]{\includegraphics[scale=0.48]{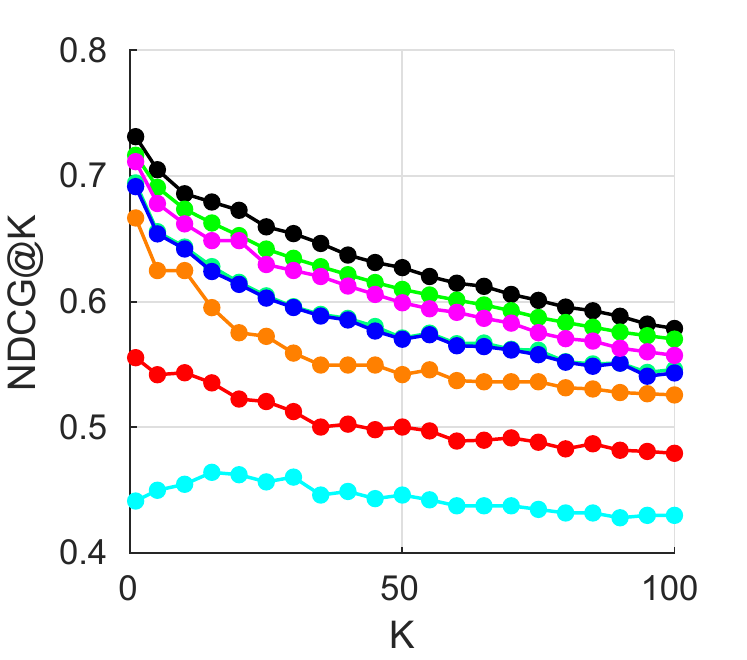}}
\subfigure[Double]{\includegraphics[scale=0.48]{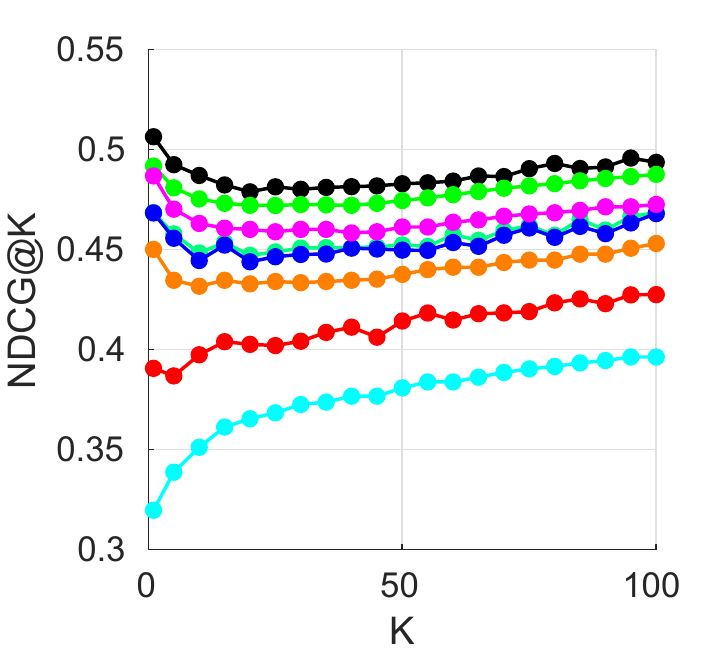}}
\subfigure[Triple]{\includegraphics[scale=0.48]{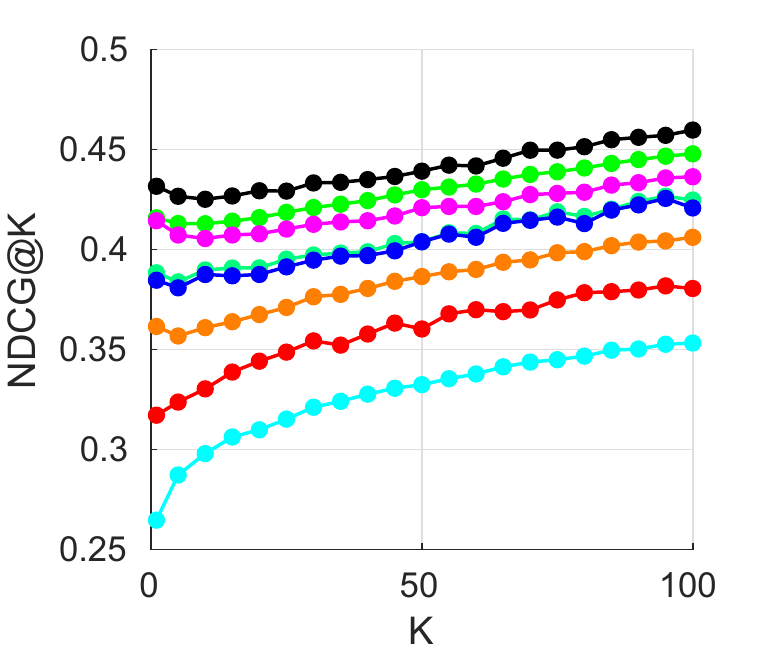}}
\caption{Ranking performance on the LFW dataset. Due to space restriction, the legend is shown in the box on the left.}
\label{fig:Rank_FaceTracer}
\vspace{-0.40cm}

\end{figure*}

\begin{figure*}
\centering
\subfigure{\includegraphics[scale=0.48]{legend.png}}
\subfigure[Single ]{\includegraphics[scale=0.43]{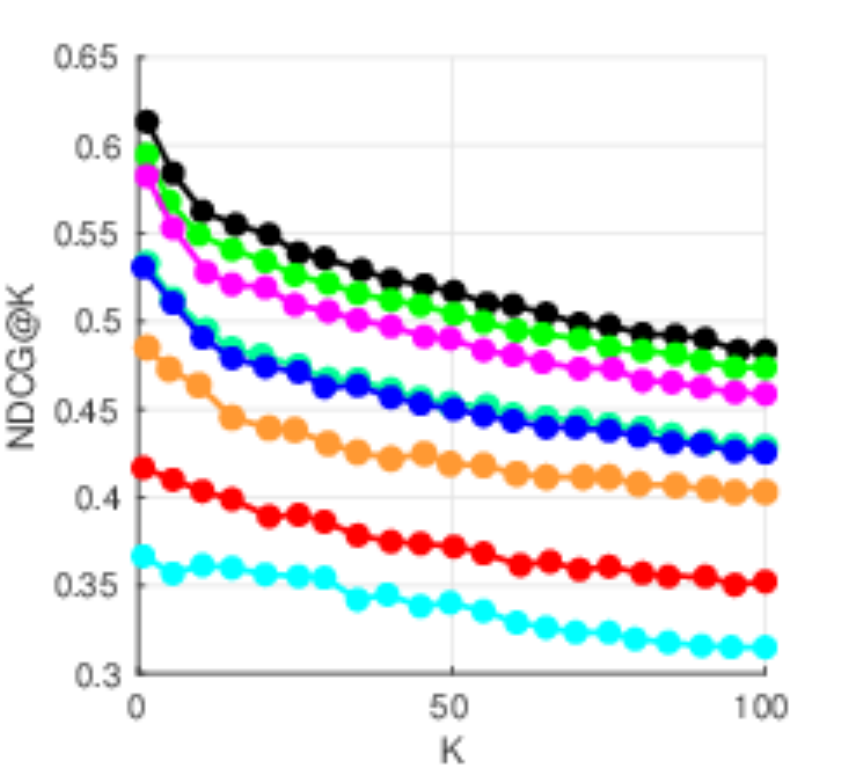}}
\subfigure[Double]{\includegraphics[scale=0.43]{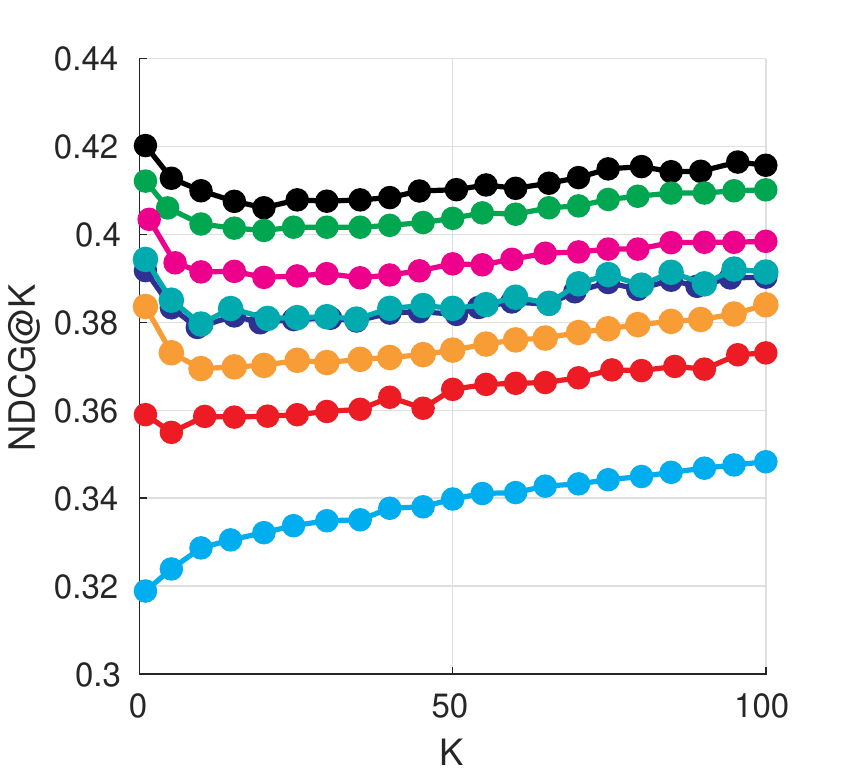}}
\subfigure[Triple]{\includegraphics[scale=0.43]{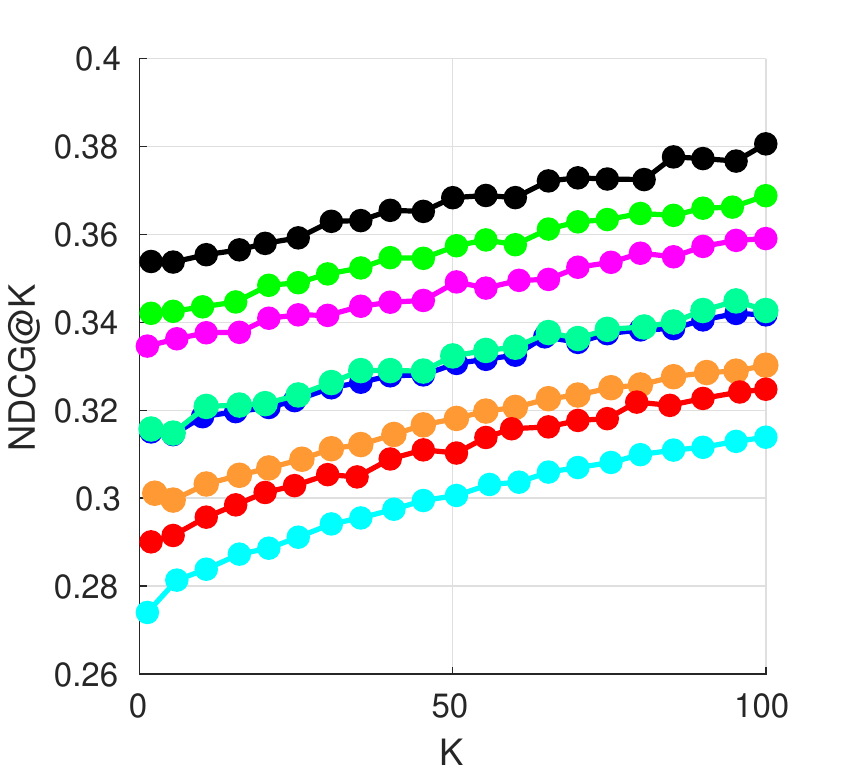}}
\caption{Ranking performance on the YouTube dataset. Due to space restriction, the legend is shown in the box on the left.}
\label{fig:Rank_YTF}
\vspace{-0.40cm}
\end{figure*}

\textbf{Effect of relation between margin $m$ and error correcting capability $e$ on the retrieval performance}: Before extensive performance evaluation of the proposed framework, it is very important to understand how the relation between error-correcting capability $e$ of the NECD used in stage 1(b) and the margin $m$ of the margin-based distance logistic loss (DLL) in stage 1(a) affects the overall retrieval performance for the proposed framework. From Sec. \ref{subsec:DNDCMH}, we know that $e \geq m$. For a given value of $m$ we need to analyze the upper limit of $e$ that will give us a reasonable retrieval performance. We use mean average-precision (MAP) as our metric to evaluate the relation between $e$ and $m$. Table \ref{table:e_vs_m} provides the MAP results for three datasets. We have considered hash code length of 63 bits with margin values as $m=3,5,6,7$. Table \ref{table:e_vs_m} also provides the different values of error correcting capability $e$ we have used for a given margin $m$. We have tried the values of $e$ in the range of $m-1$ to $m+6$ depending on the ECC word $(N,K)$ possible with hash code of 63 bits.

We can observe from Table \ref{table:e_vs_m} that for an given value of $m$, the best MAP is achieved when $e$ is equal to or slightly greater than $m$. For example, when $m=5$, we can observe that the best MAP is achieved when $e$ is either $5$ or $6$. It is interesting to note that as $e$ becomes greater and greater than $m$, the MAP starts reducing. The reason for this reduction is that as the error correcting capability $e$ increases, more and more impostors will fall within the Hamming sphere (error correcting capability), leading to more false positives and lower precision.

\begin{figure*}[t]
\centering
\subfigure[P-R curve for $\theta$]{\includegraphics[scale=0.295]{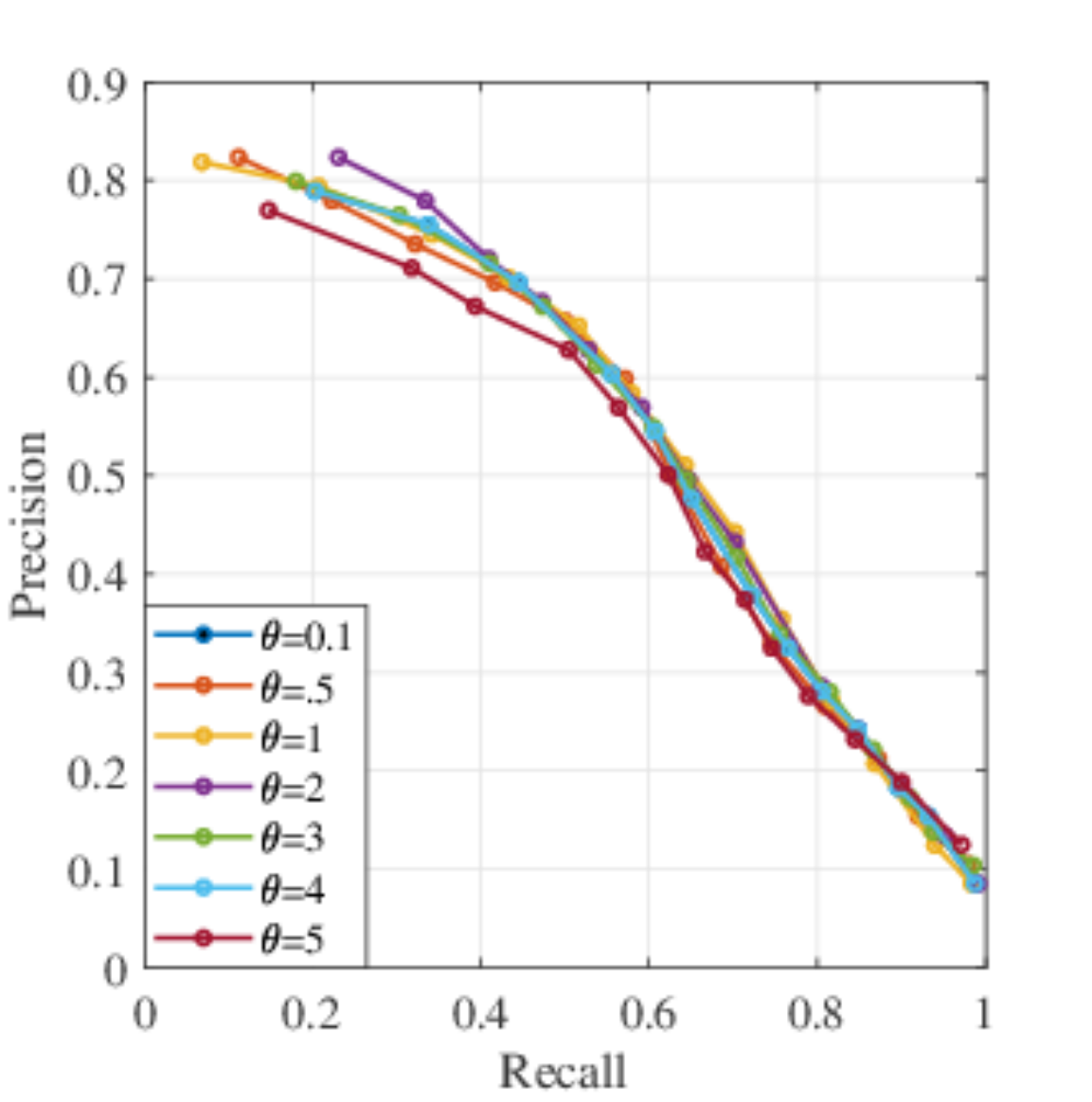}}
\subfigure[P-R curve for $\lambda$]{\includegraphics[scale=0.295]{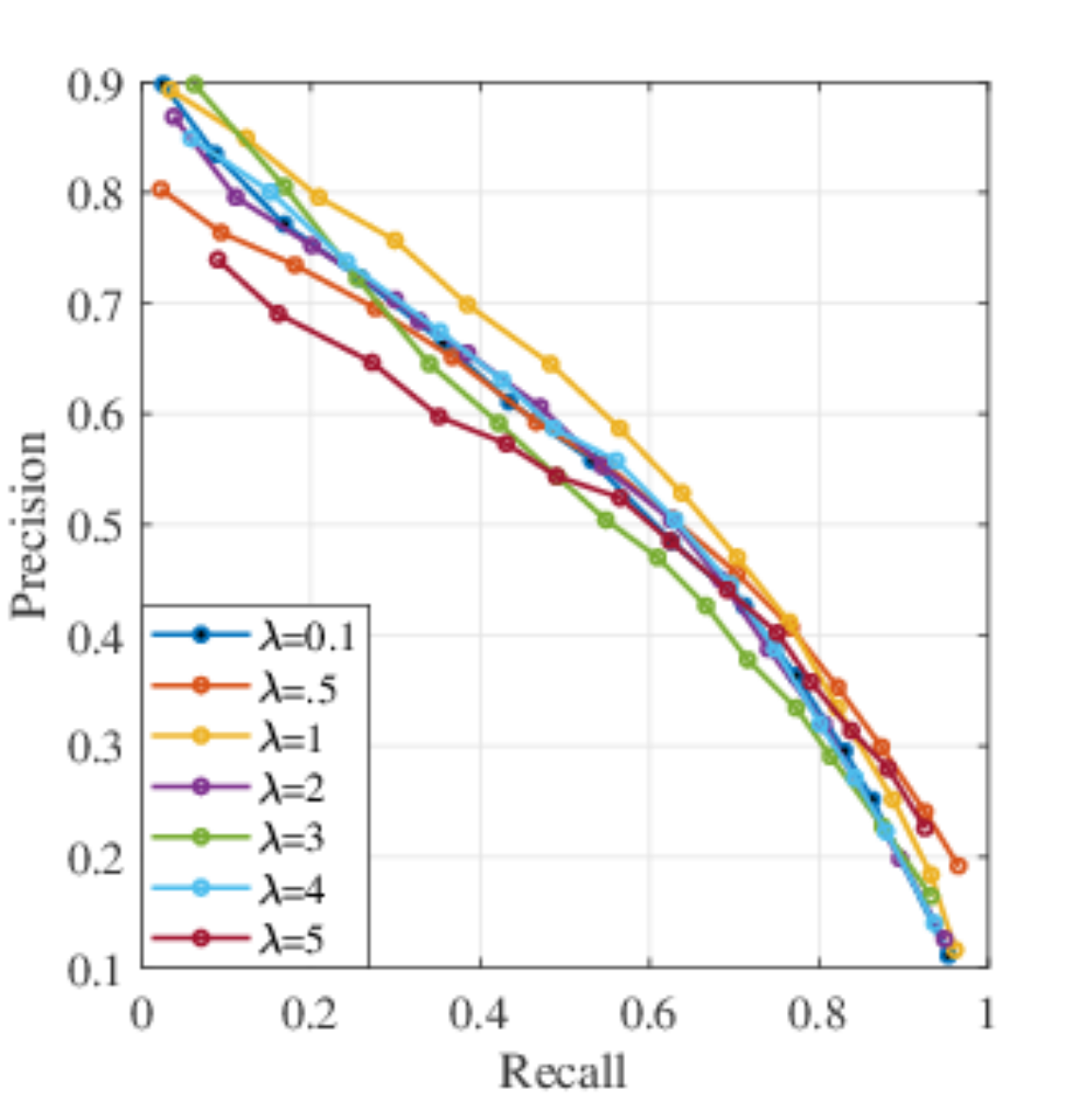}}
\subfigure[P-R curve for $\gamma$]{\includegraphics[scale=0.295]{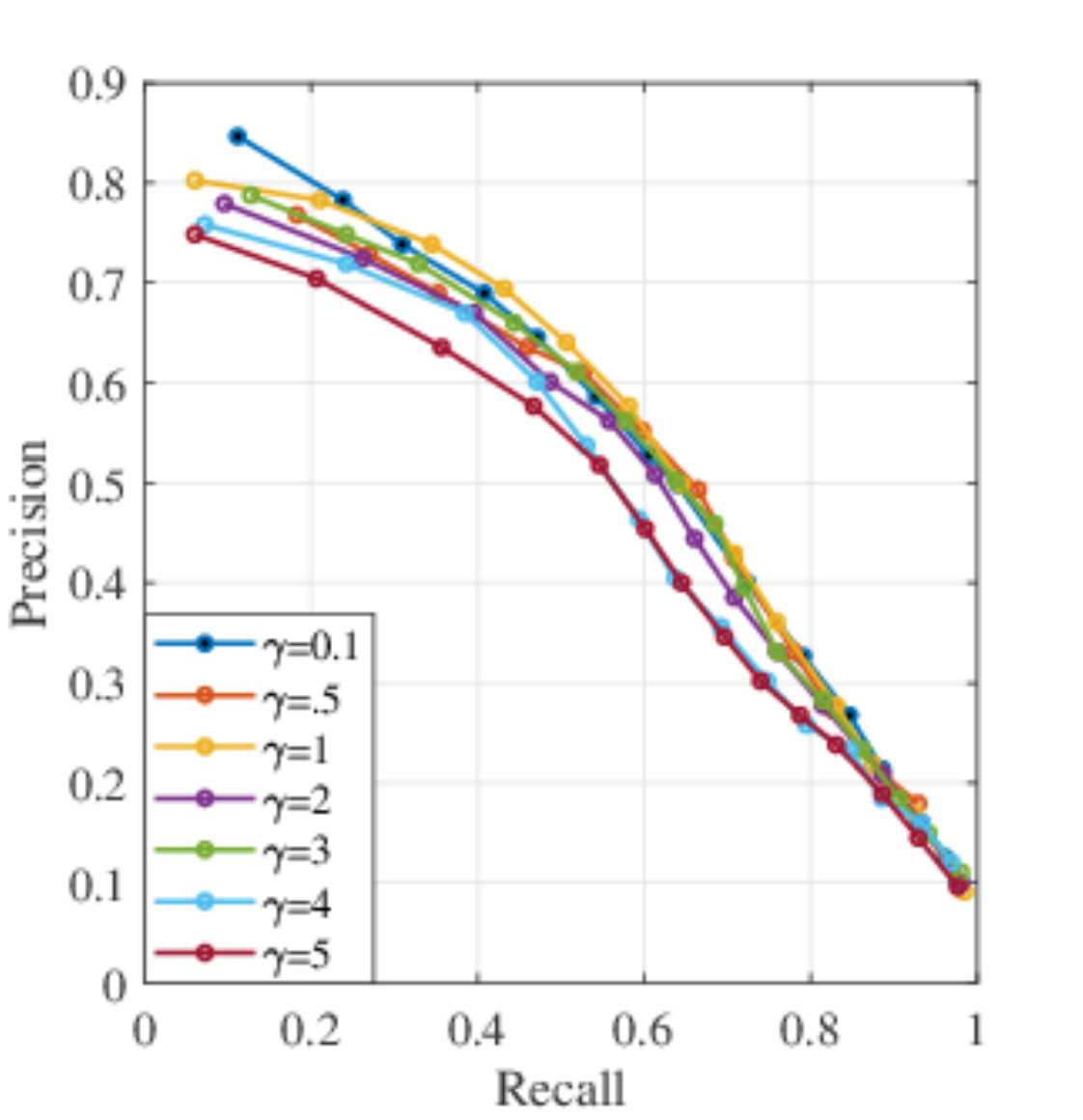}}
\caption{Influence of Hyper-parameters on P-R curves for CelebA dataset.}
\label{fig:HypParam_LFW}
\vspace{-0.40cm}

\end{figure*}

\begin{figure*}[t]
\centering
\subfigure[P-R curve for $\theta$]{\includegraphics[scale=0.295]{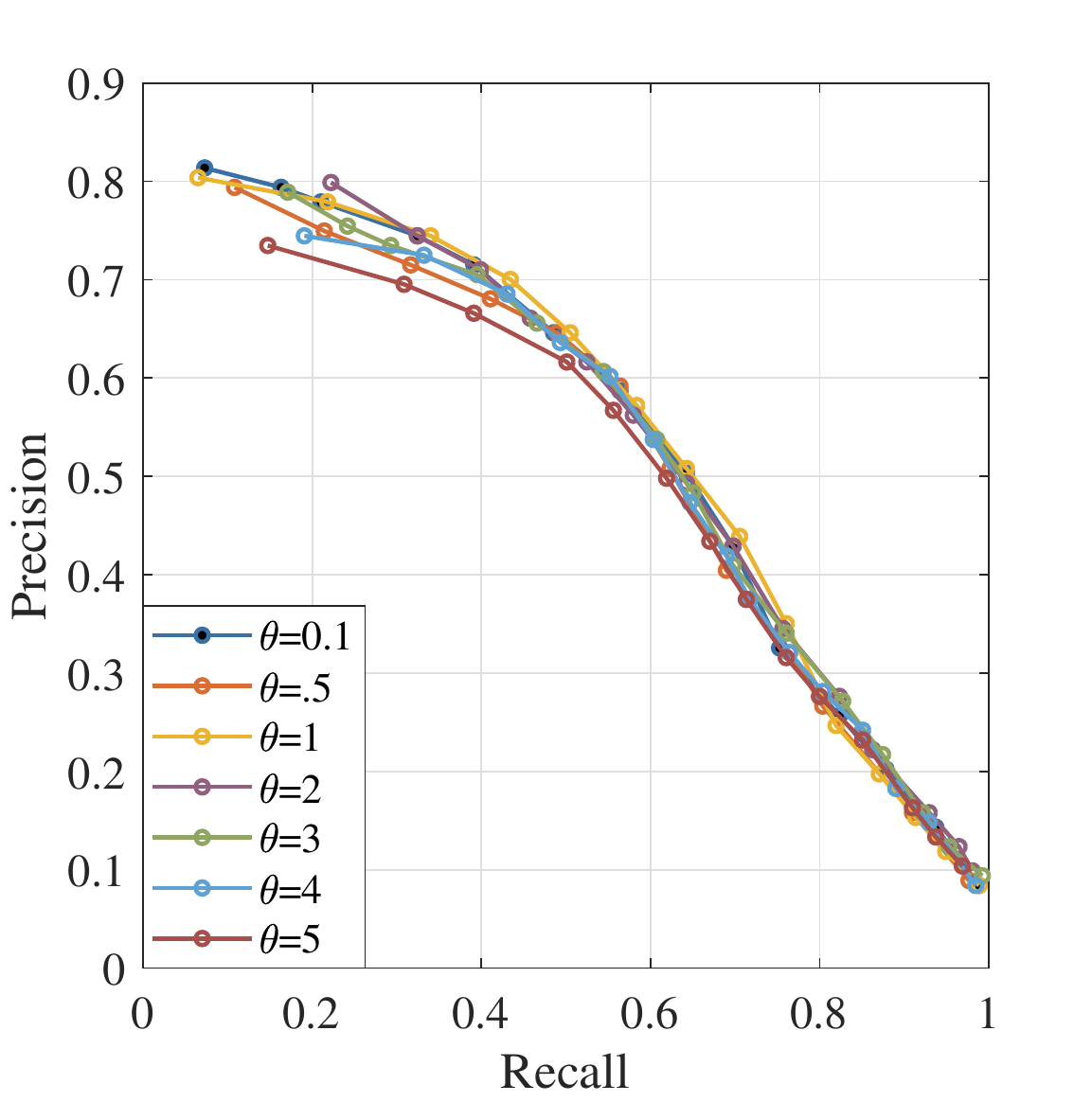}}
\subfigure[P-R curve for $\lambda$]{\includegraphics[scale=0.295]{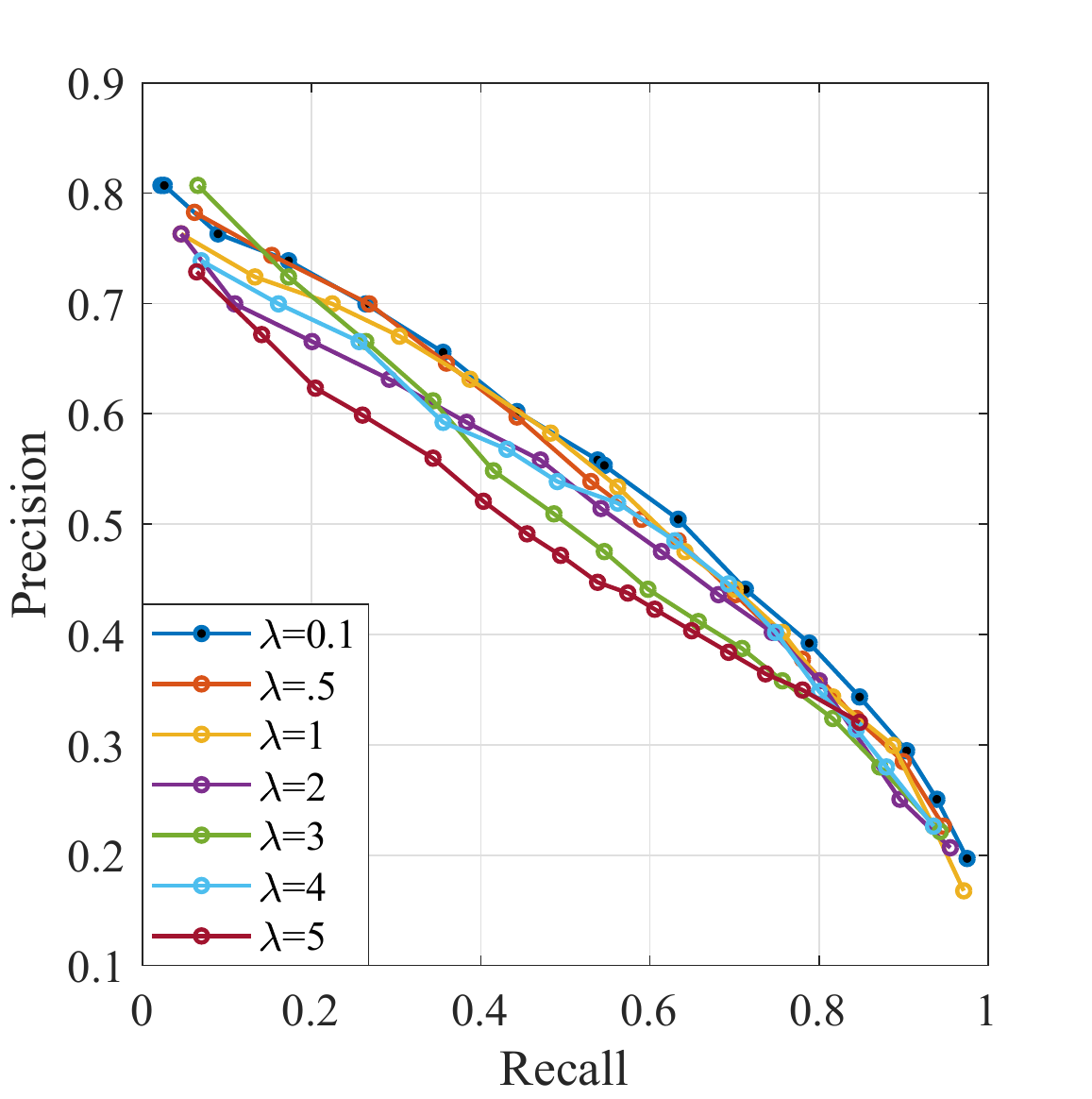}}
\subfigure[P-R curve for $\gamma$]{\includegraphics[scale=0.295]{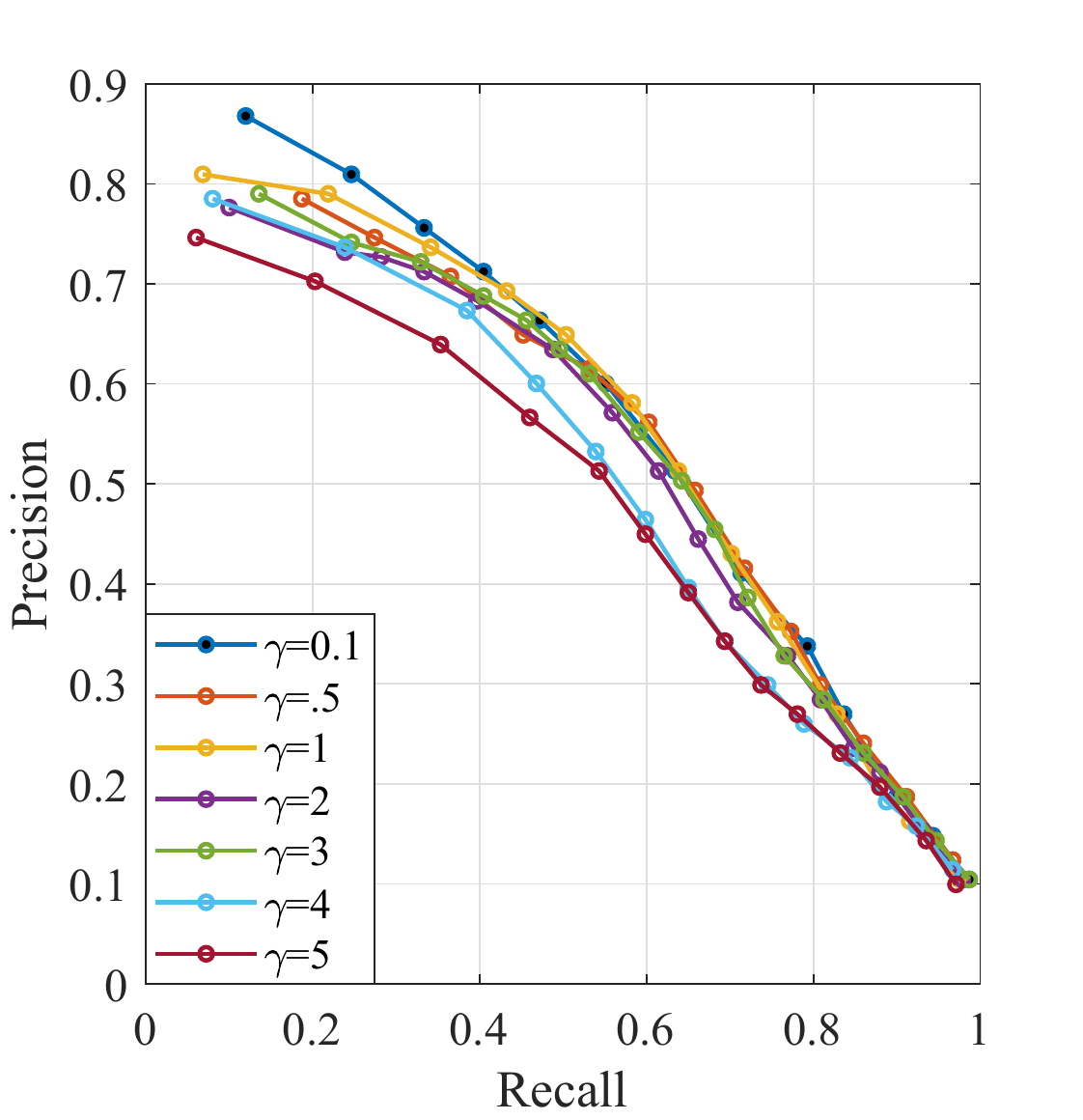}}
\caption{Influence of Hyper-parameters on P-R curves for LFW dataset.}
\label{fig:HypParam_Face_Tracer}
\vspace{-0.40cm}
\end{figure*} 
\begin{figure*}[t]
\centering
\subfigure[P-R curve for $\theta$]{\includegraphics[scale=0.41]{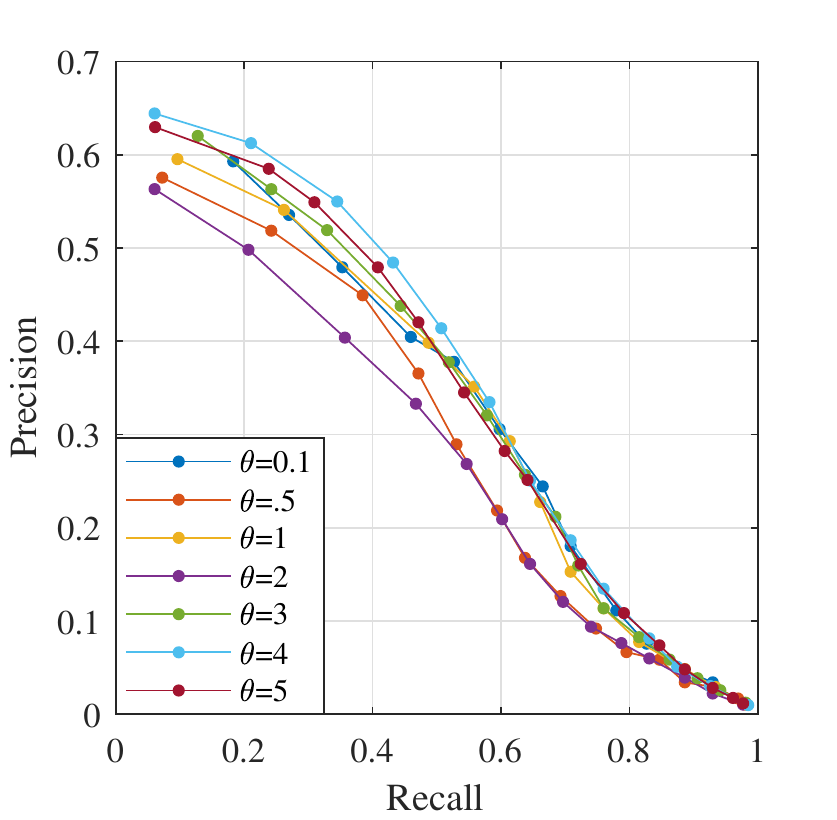}}
\subfigure[P-R curve for $\lambda$]{\includegraphics[scale=0.41]{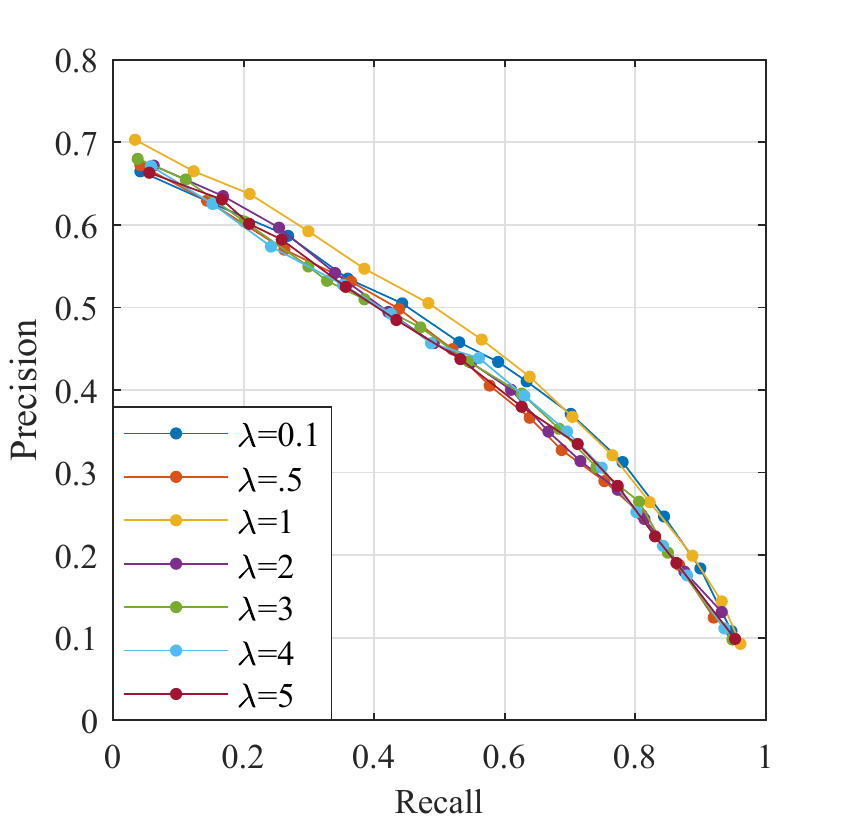}}
\subfigure[P-R curve for $\gamma$]{\includegraphics[scale=0.41]{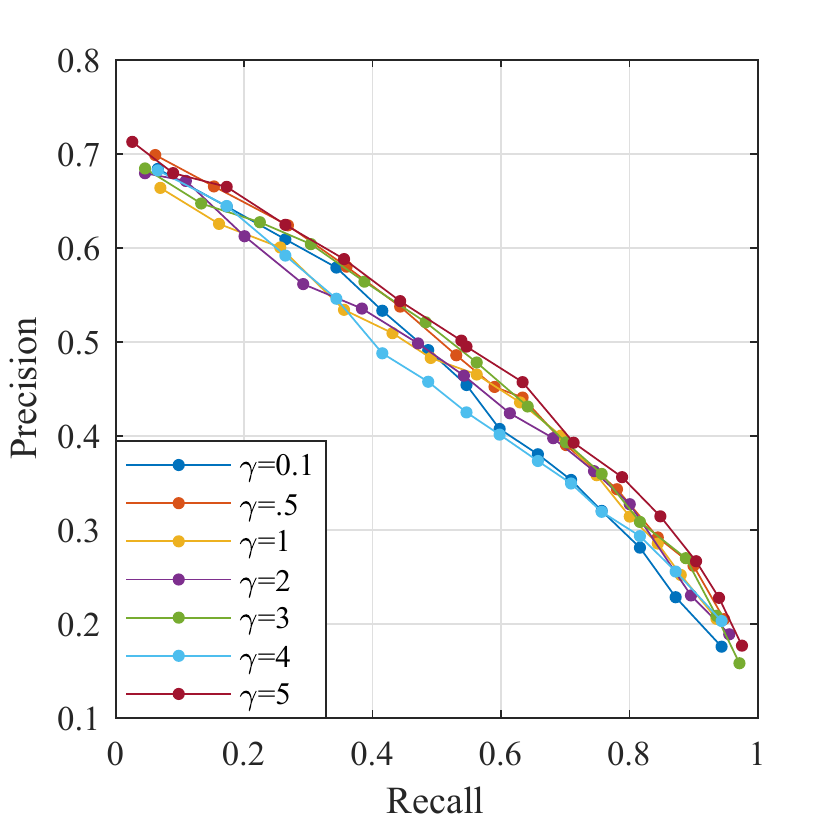}}
\caption{Influence of Hyper-parameters on P-R curves for YouTube dataset.}
\label{fig:HypParam_YTF}
\end{figure*}

\textbf{Retrieval Performance}:  MAP results using different number of attributes in the query for DNDCMH, ADCMH and other baselines for the three datasets is given in Table \ref{table:MAP_comp}. For this experiment, we have used a margin of $m=6$ for margin-based DLL. For the best results, we have trained our NECD with BCH(63,30) code, which implies that the hash code length is equal to 63 bits and the error correcting capability $e$ of the NECD is equal to 6, which implies $e=m$. We can clearly see that our DNDCMH method clearly outperforms all the other baseline methods including the ADCMH. An interesting observation is that our method ADCMH with no NECD also outperforms DCMH, which shows that the distance-based logistic loss used in our objective function in (\ref{eq:1}) is better than the negative log-likelihood loss used in DCMH. Also, the addition of NECD to ADCMH, which is our proposed DNDCMH improves the  retrieval performance and outperforms the other state-of-the-art deep cross-modal hashing methods PRDH and THN.

\begin{table}
\centering
\small
\caption{MAP comparison for DNDCMH with other baselines for the three datasets using different number of attributes in the query. The best MAP is shown in boldface.}
\scalebox{0.65}{\begin{tabular}{|c|c|c|c|c|c|c|c|c|c|}
 \hline
\multicolumn{1}{|c}{\multirow{1}{*}{}} &\multicolumn{3}{|c}{\multirow{1}{*}{CelebA}} &\multicolumn{3}{|c|}{LFW}&\multicolumn{3}{|c|}{YouTube}\\ \cline{2-10}  
\multicolumn{1}{|c}{\multirow{1}{*}{$Method$}} &\multicolumn{1}{|c}{\multirow{1}{*}{Single}}&\multicolumn{1}{|c}{\multirow{1}{*}{Double}}&\multicolumn{1}{|c}{\multirow{1}{*}{Triple}} &\multicolumn{1}{|c}{\multirow{1}{*}{Single}}&\multicolumn{1}{|c}{\multirow{1}{*}{Double}}&\multicolumn{1}{|c|}{\multirow{1}{*}{Triple}}&\multicolumn{1}{|c}{\multirow{1}{*}{Single}}&\multicolumn{1}{|c}{\multirow{1}{*}{Double}}&\multicolumn{1}{|c|}{\multirow{1}{*}{Triple}}\\ \hline \hline
TagProp &52.610 &48.494 &41.509 &51.776 &46.556 &39.790&50.393 &44.874 &39.093\\ \hline
RankBoost & 55.116& 51.354& 50.672& 53.320& 50.198& 48.259& 54.67& 52.39& 49.074\\ \hline
MARR & 61.334& 57.890& 56.098& 59.343& 54.266& 55.334&58.670& 55.712 & 53.84\\ \hline
DCMH & 67.210& 62.664& 61.170& 62.320& 61.200& 60.320& 63.886& 60.345 & 58.765\\ \hline
PRDH & 72.100& 68.550& 66.110& 66.764& 65.776& 64.37&66.111& 64.298& 61.98 \\ \hline
THN & 74.982& 71.498& 70.498& 69.907& 69.897& 67.297 & 68.564& 66.673& 63.456 \\ \hline
ADCMH & 68.437& 64.173& 63.993& 63.732& 63.171& 62.689 & 65.163& 61.932 & 60.005 \\ \hline
 DNDCMH & \textbf{79.125}& \textbf{73.167}& \textbf{72.913}& \textbf{74.873}& \textbf{71.242}& \textbf{70.643}& \textbf{71.998}& \textbf{68.01}& \textbf{65.389} \\ \hline
\end{tabular}}
\label{table:MAP_comp}

\end{table}

\textbf{Ranking Performance}: Comparison of the NDCG scores, as a function of the ranking truncation level K, using different number of attribute queries are given in Fig. \ref{fig:Rank_LFW}, Fig. \ref{fig:Rank_FaceTracer}, and Fig. \ref{fig:Rank_YTF} for the three datasets using hash code length of 63 and BCH (63,30) with $e=m=6$ for DNDCMH and ADCMH. It is clear from the figures that our approach (DNDCMH) significantly outperforms all the baseline methods for all three types of queries, at all values of K. For example, for the LFW dataset, at a truncation level of 20 (NDCG@20), for single, double and triple attribute queries, DNDCMH is respectively, $2.1\%$, $2.1\%$ and $2.0\%$ better than THN, the best deep cross-modal hashing method, and also DNDCMH is respectively, $11.2\%$, $7.3\%$ and $8.0\%$ better than MARR, the best shallow method for attribute-based image retrieval. The ranking performance using intermediate hash codes generated by only ADCMH with no NECD also outperforms the shallow methods MARR, RankBoost, and Tagprob, and also outperforms DCMH for double and triple attribute queries and is very close (may be slightly better) to DCMH performance for single-attribute queries. The better performance of DNDCMH when compared to other deep cross modal hashing method demonstrates the effectiveness of NECD in improving the performance for cross-modal retrieval. We can observe that the NDCG values for the YTF dataset for all  methods are relatively lower when compared to the other two datasets. This is due to the motion blur and high compression ratio of downloading the videos from YouTube and extracting the faces. 

\textbf{Parameter Sensitivity:} 
We explore the influence of the hyper-parameters $\theta$, $\lambda$, and $\gamma$. Fig. \ref{fig:HypParam_LFW}, Fig. \ref{fig:HypParam_Face_Tracer}, and Fig. \ref{fig:HypParam_YTF} show the precision-recall results on the three datasets, with different values of $\theta$, $\lambda$, and $\gamma$, where the code
length is 63 bits and $e=m=6$. We can see that DNDCMH is not sensitive to
$\theta$,$\lambda$, and $\gamma$ with $0.1 < \theta < 5$,$0.1 < \lambda < 5$, and $0.1 < \gamma < 5$.

\begin{table}
\centering
\small

\caption{MAP comparison for DNDCMH with ADCMH using different \# of bits.}
\scalebox{0.53}{\begin{tabular}{|c|c|c|c|c|c|c|c|c|c|c|}
 \hline
\multicolumn{1}{|c}{\multirow{2}{*}{$Task$}} &\multicolumn{1}{|c}{\multirow{2}{*}{$Method$}}   &\multicolumn{3}{|c|}{CelebA} &\multicolumn{3}{|c|}{LFW}&\multicolumn{3}{|c|}{YouTube}\\ [0.5ex] 

 \cline{3-11}
 & & 31 bits & 63 bits  & 127 bits  & 31 bits & 63 bits  & 127 bits & 31 bits & 63 bits  & 127 bits \\ \hline \hline
 \multirow{2}{*}{Single Attribute} & ADCMH & 68.158 & 68.437 & 68.513 & 63.325 & 63.732 & 63.917&66.009&65.163&65.13 \\ \cline{2-11}
 
 & DNDCMH &78.917& 79.125 & 79.565 & 74.132 & 74.873 & 75.032&72.001&71.998&71.659\\ \hline \hline
\multirow{2}{*}{Double Attribute} & ADCMH & 64.121 & 64.173 & 64.395 & 63.121 & 63.171 & 63.532&61.489&61.932&60.891 \\ \cline{2-11}
& DNDCMH &74.152& 73.167 & 75.921 & 71.112 & 71.242 & 72.056&68.34&68.01&67.901\\ \hline \hline
\multirow{2}{*}{Triple Attribute} & ADCMH & 34.577 & 63.993 & 64.634 & 62.664 & 62.689 & 62.754&60.43&60.005&59.43 \\ \cline{2-11}

& DNDCMH &72.152& 72.913 & 73.271 & 70.216 & 70.643 & 70.855&65.778&65.389&64.998\\ \hline  
\end{tabular}}

\label{table:MAP_bits}

\end{table}

\begin{figure}[t]
\centering 
\subfigure{\includegraphics[width=7.6cm, height=8.2cm]{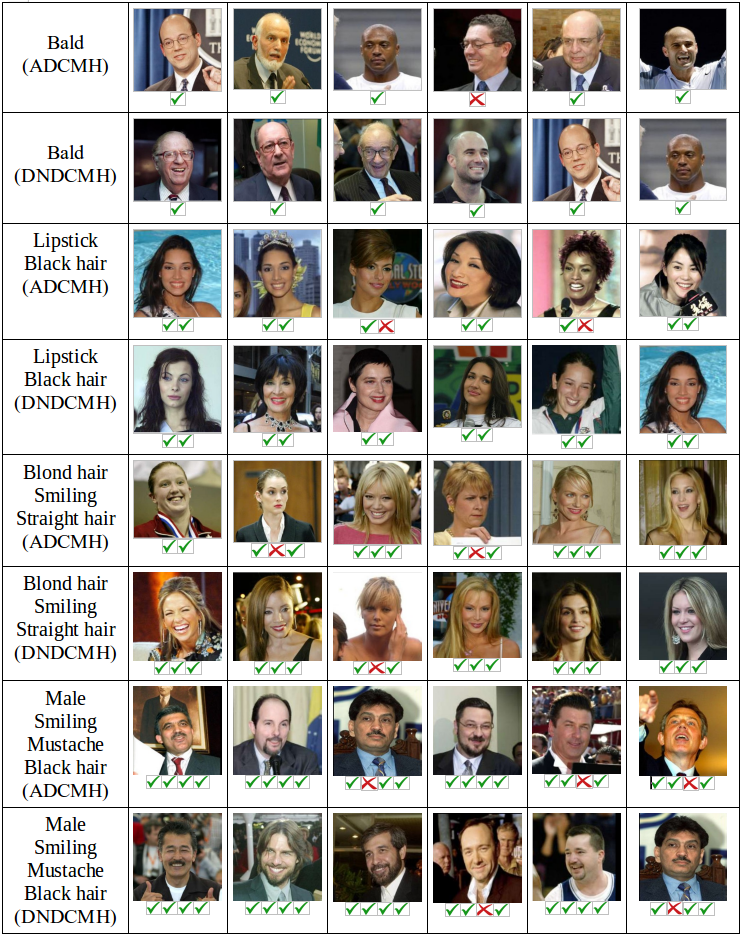}}
\caption{Qualitative results: Retrieved images using DNDCMH and ADCMH by giving different combinations of facial attributes as a query. Tick and cross symbols indicate the correct and wrong image retrieval from the testing set, respectively.}
\label{Qual}
\vspace{-0.45cm}
\end{figure}

\textbf{Effectiveness of NECD for Improving ADCMH Retrieval Performance:} To show the effectiveness of NECD combined with the ADCMH network, we conducted experiments using two different models: a) ADCMH, which indicates the case where we train the model only using stage 1(a) optimization without including NECD in the training, specifically this case considers a cross-modal hashing based on entropy, quantization and distance-based logistic loss; b) DNDCMH indicates our overall model where we include NECD and use iterative alternate optimization to correct the generated codes by ADCMH, qualitative results shown in Fig.\ref{Qual} indicate that the ADCMH retrieval performance is improved by integrating together NECD and ADCMH. 

In addition to the qualitative results, we have also compared our DNDCMH with ADCMH using MAP by varying the hash code length. Table \ref{table:MAP_bits} provides the MAP comparison for DNDCMH and ADCMH for different hash code lengths (31, 63, and 127). For hash code length of 31, 63, and 127 we have used $e=m=3$, $e=m=6$, and $e=m=11$, respectively. For ADCMH, we have used the same margin value as DNDCMH. The results in the Table show that DNDCMH gives much better results than ADCMH, which implies that additional optimization using NECD improves retrieval performance for ADCMH. Additionally, the retrieval performance does not change a lot with increase in the hash code length. Consequently, even with low storage capacity of 31 bits, high retrieval performance is achieved.   


\textbf{Effect of Number of Attributes:} From the experimental results, we can see that the retrieval or the ranking performance for DNDCMH and also ADCMH decreases with the increase in the number of facial attributes as query. This is evident from the quantitative results in Fig. \ref{Qual} and also quantitative results in Fig. \ref{fig:Rank_LFW}, and Fig. \ref{fig:Rank_FaceTracer}. The reason for this decrease is that as we increase the number of facial attributes in a query, the number of constraints to map the facial image modality into the same Hamming space as the attribute modality, also gets inflated, which leads to lower retrieval performance when compared to only a small number of facial attribute in the query.





\section{Conclusion}
In this paper, we proposed a  novel iterative two-step deep cross-modal hashing method that takes facial attributes as query and returns a list of images based on a Hamming distance similarity. In this framework, we leveraged a neural network based decoder to correct the codes generated by  the facial attribute-based deep cross-modal hashing to improve the retrieval performance. The experimental results show that the neural network decoder significantly improves the retrieval performance of the attribute-based deep cross-modal hashing network. Moreover, the results indicate that the proposed framework outperforms most of the other cross-modal hashing methods for attribute-based face image retrieval.


%


\appendices
\section{Error-Correcting Code and Decoding}\label{app:a}
\textbf{Error-Correcting Code}:  The function of the channel encoder in a digital transmission system is to transform the information sequence generated by the information source into a discrete encoded sequence called a codeword. 
The function of channel decoder is to transform  received sequence into a binary sequence called the estimated information sequence. Ideally, we want  estimated information sequence to be the same as transmitted information sequence, although the channel noise may cause some decoding errors. Error-Correcting codes are used to correct this channel noise such that the estimated information sequence is as close as possible to the transmitted information sequence.



\textbf{Linear Block Codes:} In a given code of length $n$ and $2^{k}$ codewords, if the modulo-2 sum of two codewords is also a codeword, then the given code is called a linear block code.  This implies that an $(n,k)$ linear code $\mathcal{C}$ with $2^{k}$ codewords forms a $k$-dimensional subspace of the vector space of all binary $n$-tuples over the field $GF(2)$. Based on this implication, it is possible to find $k$ linearly independent codewords $(\textbf{g}_{0},\textbf{g}_{1},\cdots,\textbf{g}_{k-1})$ in $\mathcal{C}$ such that every codeword \textbf{v} in $\mathcal{C}$ is a linear combination of these $k$ codewords
\begin{equation} \textbf{v}=u_{0}\textbf{g}_{0}+u_{1}\textbf{g}_{1}+\cdots+u_{k-1}\textbf{g}_{k-1},\end{equation} where $u_{i}=0 $ or $1$ for $0\leq{i}<k$. 


  
  Let  $\textbf{u}=(u_{0},u_{1},...,u_{k-1})$ represent the message to be encoded, the corresponding codeword can be given as
  \begin{equation}\textbf{v}=\textbf{u}\textbf{G}= u_{0}\textbf{g}_{0}+u_{1}\textbf{g}_{1}+\cdots+u_{k-1}\textbf{g}_{k-1}.\end{equation}

The rows of \textbf{G} span the linear code $\mathcal{C}$. For this reason, \textbf{G} is called a \emph{generator matrix} for the linear code $\mathcal{C}$. 



Another useful matrix associated with linear block code is the parity-check matrix, which is generally denoted by \textbf{H}.  \textbf{H} is a  $(n-k)\times n$ matrix formed by $n-k$ linearly independent rows, where any vector in the rows of \textbf{H} is orthogonal to  the row space of \textbf{G} and vice-versa. This implies that an $n$-tuple \textbf{v} is a codeword in the code $\mathcal{C}$ generated by \textbf{G} if and only if $\textbf{v}\textbf{H}^\textbf{T}=\textbf{0}$. 



\textbf{Minimum Distance and Error Correcting Capability of a Block Code:} Minimum distance is an important parameter, which determines the error-detecting and error-correcting capability of a code. Given a block code $\mathcal{C}$, the \emph{minimum distance} of the code $\mathcal{C}$, denoted by $d_{\mathsf{min}}$ is defined as the minimum Hamming distance between any two codewords of the code $\mathcal{C}$ \begin{equation}d_{\mathsf{min}}= \mathsf{min}\left\{d(\textbf{v},\textbf{z}):\textbf{v},\textbf{z} \in \mathcal{C}, \textbf{v}\neq\textbf{z}\right\}.\end{equation} 


The relation between minimum Hamming distance and the error correcting capability is given by the theorem: If $d_{\mathsf{min}}\geq 2e + 1$, the standard decoding algorithm for C can correct up to $e$ errors. If $d_{\mathsf{min}}\geq 2e + 2$,
it can correct up to $e$ errors and detect $e + 1$ errors.

\textbf{Belief Propagation Decoding:} An effective graphical representation of a parity check matrix $H$ is a Tanner graph, which provides complete representation of the code and also helps to describe the decoding algorithm. Tanner graph is a bipartite graph, where the nodes of the graph are separated into two different sets and edges are only connecting nodes from one set to the other set. The two different sets of nodes in a Tanner graph are called variable nodes (v-nodes) and check nodes (c-nodes).

\begin{figure}[h]
\centering
\includegraphics[width=6cm]{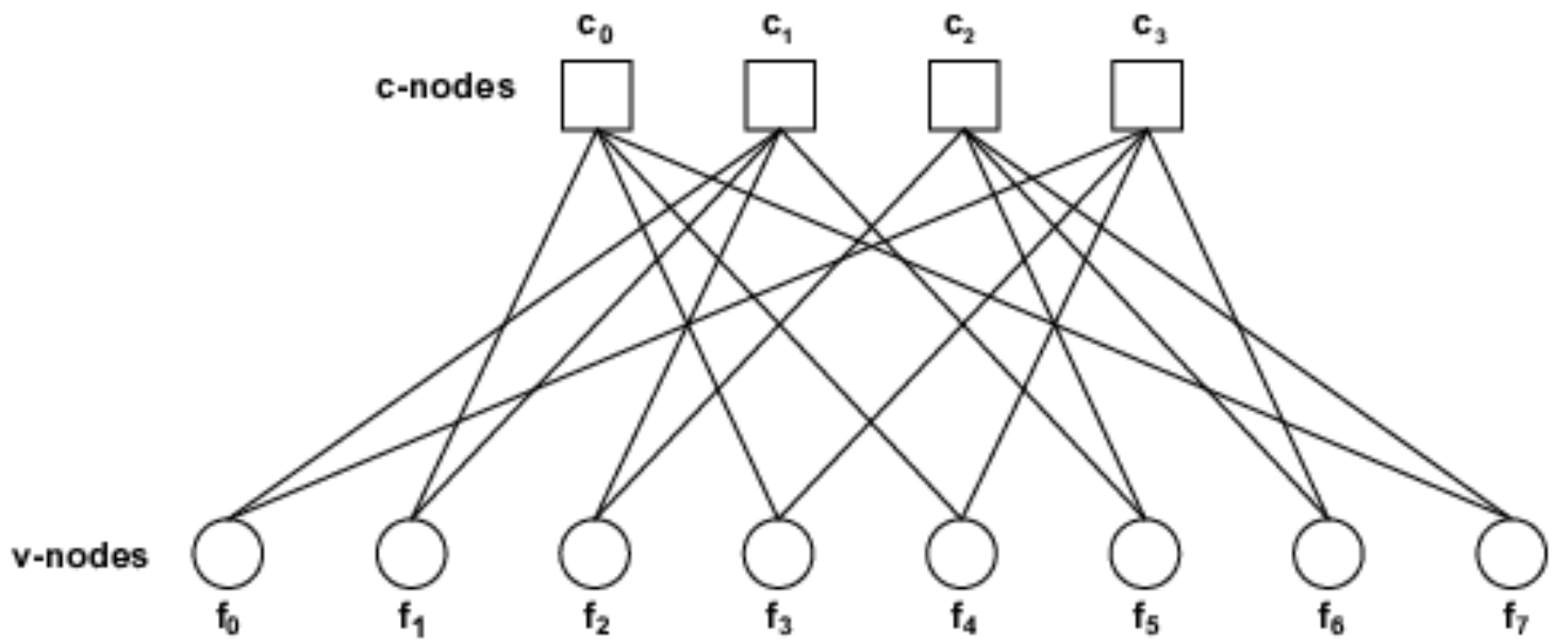}
\caption{Tanner Graph for the parity check matrix shown in (\ref{eqn:H}).}\label{fig:tanner}

\end{figure}

Fig. \ref{fig:tanner} shows the Tanner graph of a code whose parity check matrix is given as:

\begin{equation}
H=\begin{bmatrix} 
0 & 1 & 0 & 1 & 1 & 0 & 0 & 1 \\
1 & 1 & 1 & 0 & 0 & 1 & 0 & 0\\
0 & 0 & 1 & 0 & 0 & 1 & 1 & 1\\
1 & 0 & 0 & 1 & 1 & 0 & 1 & 0
\end{bmatrix}.\label{eqn:H}
\end{equation}

Building a Tanner graph is very straight forward. The check nodes (c-nodes) represent the number of parity bits (\# of rows in \textbf{H}) and the variable nodes (v-nodes) represent the number of bits in a codeword (\# of columns in \textbf{H}). Check node $c_j$ is connected to variable node $f_i$ if the element $h_{ji}$ of parity check matrix H is a 1.

The renowned belief propagation (BP) decoder can be described clearly using a Tanner graph. Assume a binary codeword
$(x_1,x_2,\cdots\cdots,x_n)$ is transmitted over an additive white Gaussian noise (AWGN) channel and the received symbol is $(y_1,y_2,\cdots\cdots,y_n)$. Let $y=x+r$ where $r$ is a noise added by the channel. The $n$ code bits must satisfy all parity checks and this will be used to compute the posterior probability $Pr(x_i/S_i,y)$ and $S_i$ is the event that all parity checks associated with
$x_i$ have been satisfied.

The BP algorithm is an iterative algorithm based on Tanner graph and is based on computation of  $Pr(x_i=1/y)$. The steps in this iterative algorithm are:

\begin{enumerate}
    \item In first step, each v-node $f_i$ processes its input message received from the channel $y_i$ and passes its resulting output message to neighboring c-node because in first pass there is no other information to be passed. 
\begin{figure}[t]
\centering
\subfigure[c-node to v-node]{\includegraphics[scale=0.48]{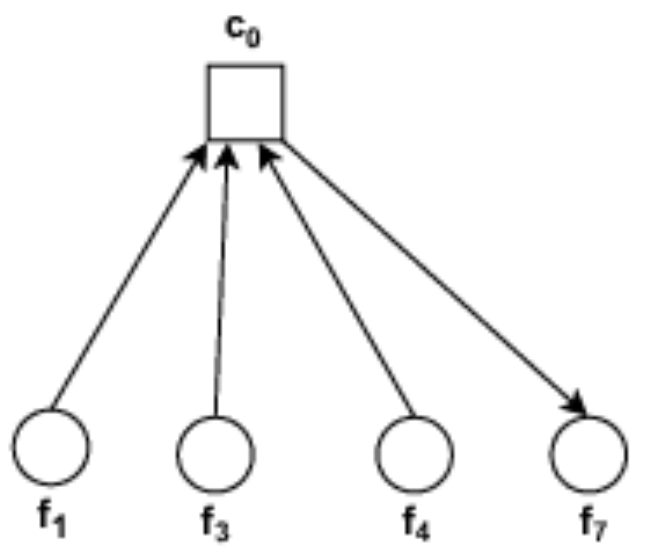}\label{fig:sub_tanner_a}}
\subfigure[v-node to c-node]{\includegraphics[scale=0.48]{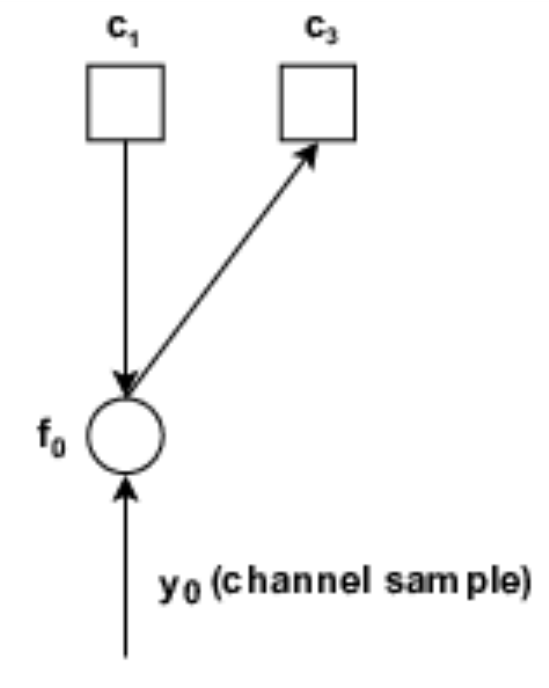}\label{fig:sub_tanner_b}}
\caption{Tanner sub-graphs showing the transfer of information from c-node to v-node and vice-versa.}
\label{fig:tanner_sub_graph}
\vspace{-0.35cm}
\end{figure}

    \item In the second step, the c-node gets the input messages passed from the v-nodes and checks whether the parity check equations are satisfied. And then passes its resulting output messages to all the connected v-nodes $f_i$  using the incoming messages from all other v-nodes, but excluding the information from $f_i$. This can be seen in Fig. \ref{fig:sub_tanner_a}. Note that the information passed to v-node $f_7$ is all the information available to c-node $c_0$ from the neighboring v-nodes, excluding v-node $f_7$. The information passed is $Pr (\text{check equation is satisfied}|\text{input messages}) $ i.e., the probability that there is an even number of 1’s among the variable nodes except the node $f_i$.  Such extrinsic information is computed for each connected c-node/v-node pair in this step.
    \item In the third step, each v-node processes its input message and passes its resulting output message to neighboring c-nodes using channel samples and incoming messages from all other c-nodes connected to v-node $f_i$, except the c-node $c_j$. This is shown in Fig. \ref{fig:sub_tanner_b}. The information passed is
$Pr(x_i=b|\text{input messages})$, where $b\in {0,1}$. 




Additionally, at this point the v-nodes also update their current estimation $\hat{y_i}$
of their variable $y_i$ by using the channel information and messages from the neighboring c-nodes, without excluding any c-node information. If the current estimated codeword fulfills now the parity check
equations the algorithm terminates. Otherwise termination is
ensured through a maximum number of iterations.
\item Go to Step 2.
\end{enumerate}





{\small
\bibliographystyle{IEEEtran}
\bibliography{egbib}
}

\vspace{-0.65cm}
\begin{IEEEbiography}[{\includegraphics[width=1in,height=1.25in,clip,keepaspectratio]{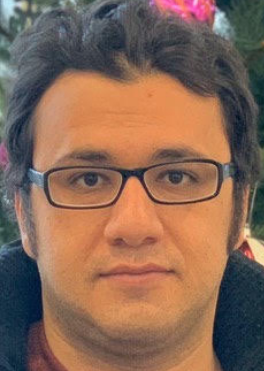}}]
{Fariborz Taherkhani} is currently a Ph.D candidate at Lane Department of Computer Science and Electrical Engineering, West Virginia University, Morgantown,  WV,  USA. He received his BSc. degree in computer engineering from National University of Iran, Tehran, Iran, and his MSc. degree in computer engineering from Sharif University of Technology, Tehran, Iran. His research interests include machine learning, computer vision, biometrics, and image retrieval. 
\end{IEEEbiography}

\vspace{-0.75cm}
\begin{IEEEbiography}[{\includegraphics[width=1in,height=1.50in,clip,keepaspectratio]{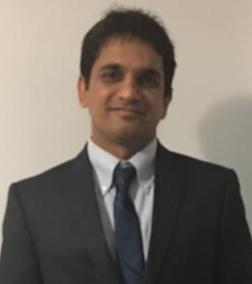}}]{Veeru Talreja} is a Ph.D. candidate at West Virginia  University,  Morgantown,  WV, USA.  He received the  M.S.E.E. degree from  West Virginia  University and the B.Engg. degree from Osmania University, Hyderabad, India. From 2010 to 2013 he worked as a Geospatial Software Developer with West Virginia University Research corporation. His  research  interests include applied machine learning, deep learning, coding theory, multimodal biometric recognition and security, and image retrieval.  
\end{IEEEbiography}

 \vspace{-0.65cm}
\begin{IEEEbiography}[{\includegraphics[width=1in,height=1.25in,clip,keepaspectratio]{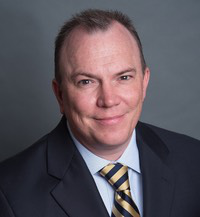}}]{Matthew  C.  Valenti} (M’92  -  SM’07  -  F’18)  received the M.S.E.E. degree from the Johns Hopkins University, Baltimore, MD, USA, and B.S.E.E. and Ph.D. degrees from Virginia Tech, Blacksburg, VA,USA.  He  has  been  a  Faculty  Member  with  West Virginia University since 1999, where he is currently a Professor and the Director of the Center for Identification Technology Research. His research interests are  in  wireless  communications,  cloud  computing, and biometric identification. He is the recipient of the 2019 MILCOM Award for Sustained Technical Achievement. He is active in the organization and oversight of several ComSoc sponsored IEEE conferences, including MILCOM, ICC, and Globecom. He was Chair of the  ComSoc Communication  Theory  Technical  committee  from  2015-2016, was  TPC  chair  for  MILCOM’17,  is  Chair  of  the  Globecom/ICC  Technical Content  (GITC)  Committee  (2018-2019), and  is  TPC  co-chair  for  ICC’21 (Montreal).  He  was  an  Electronics  Engineer  with  the  U.S.  Naval  Research Laboratory, Washington, DC, USA. Dr. Valenti is registered as a Professional Engineer in the state of West Virginia
\end{IEEEbiography}
\vspace{-0.65cm}
\begin{IEEEbiography}[{\includegraphics[width=1.25in,height=1.25in,clip,keepaspectratio]{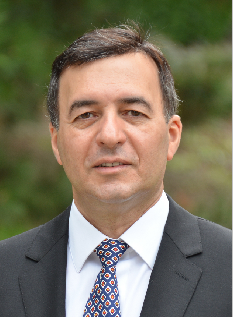}}]{Nasser M. Nasrabadi} (S’80 – M’84 – SM’92 – F’01) received the B.Sc. (Eng.) and Ph.D. degrees in electrical engineering from the Imperial College of Science and Technology, University of London, London, U.K., in 1980 and 1984, respectively. In 1984, he was with IBM, U.K., as a Senior Programmer. From 1985 to 1986, he was with the Philips Research Laboratory, New York, NY, USA, as a member of the Technical Staff. From 1986 to 1991, he was an Assistant Professor with the Department of Electrical Engineering, Worcester Polytechnic Institute, Worcester, MA, USA. From 1991 to 1996, he was an Associate Professor with the Department of Electrical and Computer Engineering, State University of New York at Buffalo, Buffalo, NY, USA. From 1996 to 2015, he was a Senior Research Scientist with the U.S. Army Research Laboratory. Since 2015, he has been a Professor with the Lane Department of Computer Science and Electrical Engineering. His current research interests are in image processing, computer vision, biometrics, statistical machine learning theory, sparsity, robotics, and neural networks applications to image processing. He is a fellow of ARL and SPIE and has served as an Associate Editor for the IEEE Transactions on Image Processing, the IEEE Transactions on Circuits, Systems and Video Technology, and the IEEE Transactions on Neural Networks.
\end{IEEEbiography}

\end{document}